\documentclass[a4paper,12pt]{article}
\usepackage{graphicx} 
\usepackage[T1]{fontenc}
\usepackage{parskip}
\usepackage{latexsym,amsmath,amssymb}
\usepackage{amsthm} 
\usepackage{times}
\usepackage{zi4}
\usepackage[a4paper,left=2.5cm,right=2.5cm,top=2.5cm,bottom=2.5cm]{geometry}
\usepackage{hyperref}
\usepackage{float}
\usepackage{subfigure}
\usepackage{enumerate}
\usepackage{bm,bbm}
\usepackage{comment}
\usepackage{listings}

\usepackage{multirow,booktabs,setspace,caption,subcaption,float,threeparttable,authblk}
\usepackage{tikz}
\usepackage[ruled,vlined]{algorithm2e}

\usepackage{natbib}
\theoremstyle{plain} 
\newtheorem{theorem}{Theorem}[section] 

\theoremstyle{remark} 

\theoremstyle{definition} 

\title{Generative Adversarial Networks for High-Dimensional Item Factor Analysis:\\ A Deep Adversarial Learning Algorithm}
\def\correspondingauthor{\footnote{Corresponding author: Feng Ji, University of Toronto, \texttt{f.ji@utoronto.ca}}}
\author[1]{Nanyu Luo}
\author[1,\correspondingauthor]{Feng Ji}
\affil[1]{University of Toronto}
\date{November 2024}

\begin{document}

\maketitle

\begin{abstract}
Advances in deep learning and representation learning have transformed item factor analysis (IFA) in the item response theory (IRT) literature by enabling more efficient and accurate parameter estimation. Variational Autoencoders (VAEs) have been one of the most impactful techniques in modeling high-dimensional latent variables in this context. However, the limited expressiveness of the inference model based on traditional VAEs can still hinder the estimation performance. We introduce Adversarial Variational Bayes (AVB) algorithms as an improvement to VAEs for IFA with improved flexibility and accuracy. By bridging the strengths of VAEs and Generative Adversarial Networks (GANs), AVB incorporates an auxiliary discriminator network to reframe the estimation process as a two-player adversarial game and removes the restrictive assumption of standard normal distributions in the inference model. Theoretically, AVB can achieve similar or higher likelihood compared to VAEs. A further enhanced algorithm, Importance-weighted Adversarial Variational Bayes (IWAVB) is proposed and compared with Importance-weighted Autoencoders (IWAE). In an exploratory analysis of empirical data, IWAVB demonstrated superior expressiveness by achieving a higher likelihood compared to IWAE. In confirmatory analysis with simulated data, IWAVB achieved similar mean-square error results to IWAE while consistently achieving higher likelihoods. When latent variables followed a multimodal distribution, IWAVB outperformed IWAE. With its innovative use of GANs, IWAVB is shown to have the potential to extend IFA to handle large-scale data, facilitating the potential integration of psychometrics and multimodal data analysis.
\\~\\


\end{abstract}
\section{Introduction}
Item Factor Analysis (IFA; \citealp[e.g.,][]{bock1988full, wirth2007item}), is a powerful statistical framework used to model and measure latent variables. The latent variables can be mathematical ability \citep{rutkowski2013handbook}, mental disorder \citep{stochl2012mokken}, or personality traits \citep{balsis2017item} that influence individuals' responses to various items, such as test or survey questions. IFA aims to uncover these hidden factors by analyzing patterns in the responses, allowing researchers to understand the relationships between the latent variables and the items. Understanding these latent structures is crucial across fields such as psychology and education for accurate measurement of human abilities and characteristics \citep{wirth2007item,chen2021item}.

Psychological and educational research now frequently involves diverse data types, including large-scale assessment, high-volume data, and process data involving unstructured and textual responses \citep[e.g.,][]{weston2015undergraduate, zhang2023accurate, ma2024note}. Traditional estimation algorithms often struggle to efficiently process and analyze these data, leading to unstable model estimates, long computation times, and excessive memory usage. Deep Learning (DL) literature has much to offer to handle the above challenges. If we view IFA from the lens of generalized latent variable models \citep[][]{muthen2002beyond, rabe2008multilevel}, representation learning \citep{bengio2013representation} can help model and capture intricate latent patterns from observed data. Unlike traditional IFA estimation methods, which often rely on strict parametric and distributional assumptions, representation learning leverages powerful learning algorithms and expressive models such as neural networks to generate richer and highly flexible representations when modeling data. This improves the modeling and detection of subtle patterns for high-dimensional latent variables that classical methods might underfit. Therefore, by integrating representation learning with IFA, researchers can achieve a more comprehensive understanding of latent traits.

One of the representation techniques that is being actively applied in IFA \citep{hui2017variational,jeon2017variational,cho2021gaussian} is variants based on Variational Autoencoders (VAEs; \citealp{curi2019interpretable, wu2020variational}). VAEs consist of two main components: an encoder and a decoder. The encoder network compresses the high-dimensional input data into a lower-dimensional latent space, effectively capturing the essential features that define the data's underlying structure. Conversely, the decoder network reconstructs the original data from these latent representations, ensuring that the compressed information retains the critical characteristics. \citet{urban2021deep} further incorporated an importance-weighted (IW) approach \citep{burda2015importance} to improve the estimation performance of VAEs in IFA. These advances have advantage over the traditional  techniques such as Gauss-Hermite quadrature or the EM algorithm \citep{darrell1970fitting,bock1981marginal,cai2010high} for large sample size and models with high dimensional latent variables.

Despite the significant advancements brought by VAEs in IFA, conventional VAEs may inadequately capture the true posterior distribution due to insufficiently expressive inference models \citep{mescheder2017adversarial}. Specifically, VAEs might focus on the modes of the prior distribution and fails to represent the latent variables in some local regions \citep{makhzani2015adversarial}. Consequently, enhancing the expressiveness of the approximate posterior is imperative for improved performance \citep{cremer2018inference}. 

\begin{figure}[!htb]
    \centering
    \includegraphics[width=0.8\linewidth]{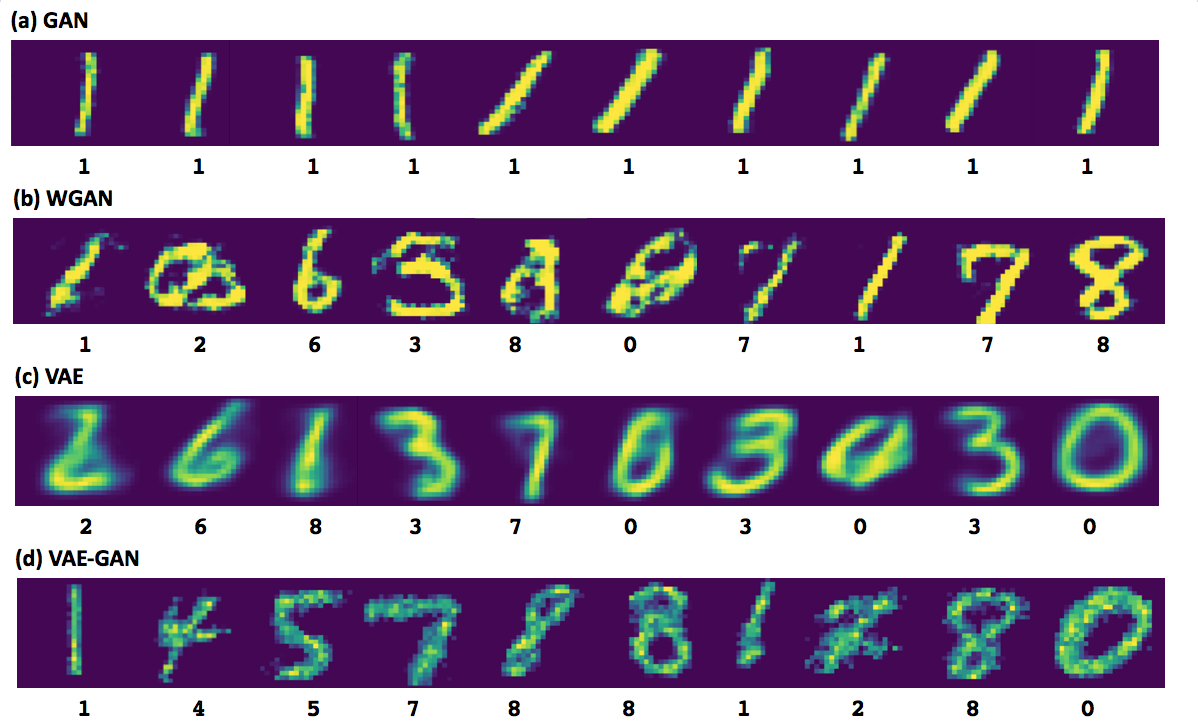}
    \caption{Ten handwritten images sampled from model (a) GAN, (b) WGAN, (c) VAE, and (d) VAE-GAN. Adapted from \cite{mi2018probe}.}
    \label{fig:image_quality}
\end{figure}

An alternative representation paradigm, Generative Adversarial Networks \citep[GANs,][]{goodfellow2014generative}, a celebrated framework for estimating generative models in deep learning and computer vision, offer greater flexibility in modeling arbitrary probability densities and can contribute to the expressiveness. Multiple variants of GANs have achieved remarkable success in complex tasks such as image generation \citep[e.g.,][]{radford2015unsupervised, arjovsky2017wasserstein, zhu2017unpaired}. Among these models, the combination of GANs and VAEs can be traced back to Adversarial Autoencoders (AAEs; \citealp{makhzani2015adversarial}) and the VAE-GAN model \citep{larsen2016autoencoding}. \textbf{Figure \ref{fig:image_quality}} provides an intuitive visual example demonstrating how integrating GANs and VAEs can lead to better representation learning. The figure compares ten handwritten images sampled from four models: (a) GAN, (b) WGAN \citep[a GAN variant,][]{arjovsky2017wasserstein}, (c) VAE, and (d) VAE-GAN. While GAN generated only the digit 1\footnote{{The generator's output of only `1's indicates mode collapse, a common failure mode in vanilla GAN training. The standard GAN objective encourages the network to repeatedly generate identical samples that incur minimal loss, rather than to explore diverse outputs that might be more heavily penalized \citep{mi2018probe}.}}, WGAN exhibited unstable performance, and the images produced by VAE appeared blurry. In contrast, VAE-GAN generated relatively clear results {for most of digits}. However, vanilla GANs, VAE-GANs, and AAEs rely solely on adversarial loss, formulated as a minimax game between networks to deceive a discriminator. As a result, they are not motivated from a maximum-likelihood perspective and are therefore not directly applicable for improving Marginal Maximum Likelihood (MML) estimation in IFA.

\begin{figure}[!htb]
    \centering
    \subfigure[VAE]{
        \includegraphics[width=0.35\textwidth]{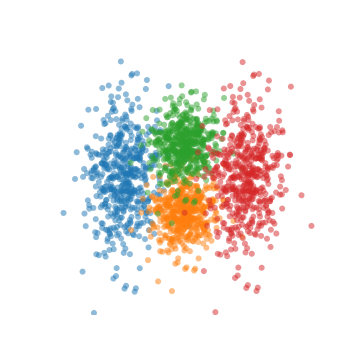}
        \label{fig:toy_comp_vae}}
    \subfigure[AVB]{
        \includegraphics[width=0.35\textwidth]{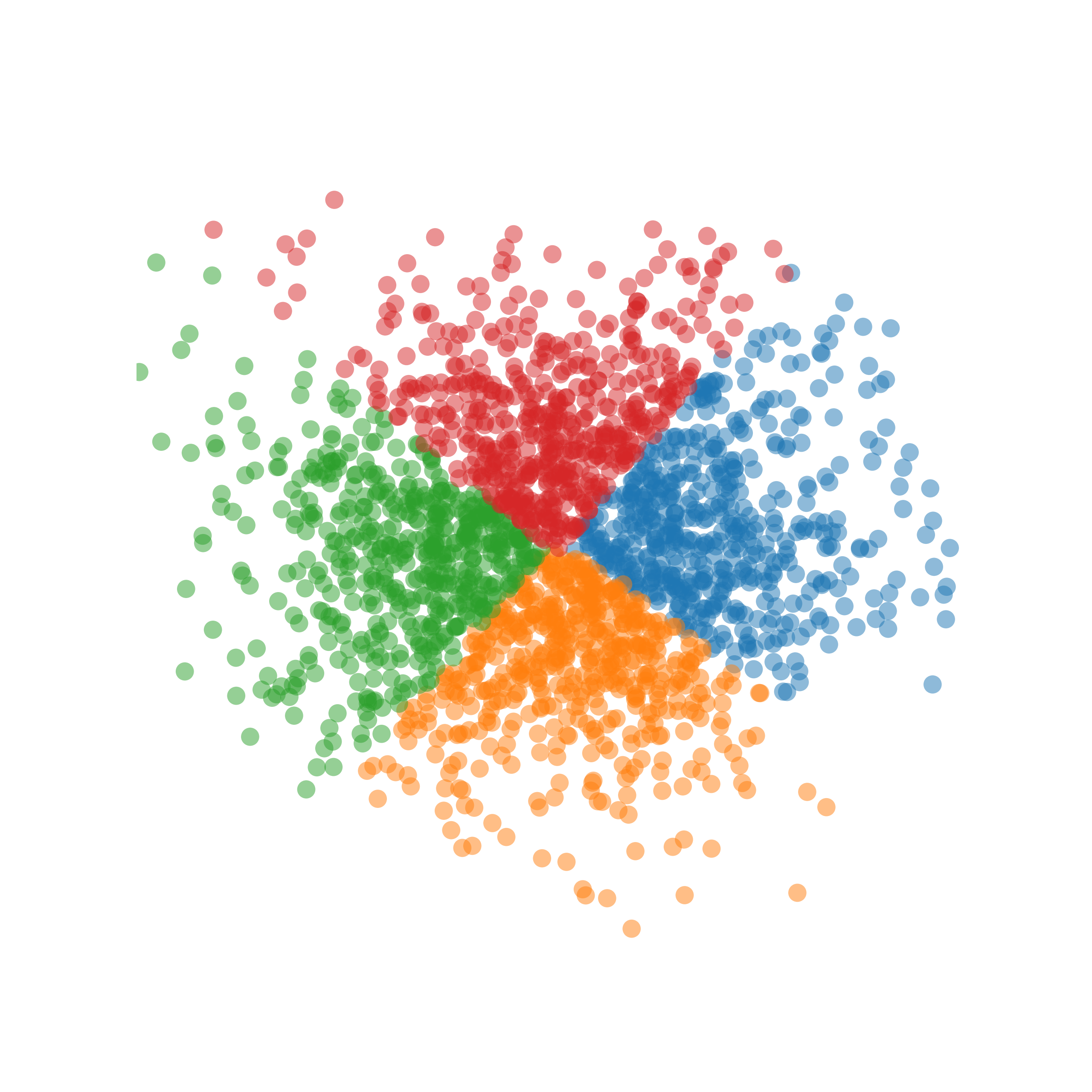}
        \label{fig:toy_comp_avb}}

    \caption{Distribution of latent variables for VAE and AVB trained on a simple synthetic dataset containing samples from 4 different labels.}
    \label{fig:toy_comp}
\end{figure}

Therefore, this paper focuses on an improved approach based on VAEs, Adversarial Variational Bayes  \citep[AVB,][]{mescheder2017adversarial}, which has a flexible and general approach to construct inference models. AVB improves the traditional VAE framework by integrating elements from GANs, specifically by introducing an auxiliary discriminator network. This discriminator engages in an adversarial training process with the encoder to help the inference model produce more accurate and expressive latent variable estimates. By combining the strengths of both VAEs and GANs, AVB removes the restrictive assumption of standard normal distributions in the inference model and enables effective exploration of complex, multimodal latent spaces. Theoretical and empirical evidence has shown that AVB can achieve higher log-likelihood, lower reconstruction error and more precise approximation of the posterior \citep{mescheder2017adversarial}. {For example, we consider a simple dataset containing samples from 4 different labels whose latent codes are uniformly distributed. \textbf{Figure \ref{fig:toy_comp}} compares the recovered latent distribution: the unallocated regions (``empty space'') and overlap between regions in VAE's result can yield ambiguous latent codes and compromise both inference precision and the quality of generated samples, but AVB produces a posterior with well-defined boundaries around each region.} This approach not only improves the flexibility and accuracy of MML estimation but also provides more potential for better handling more diverse and high-dimensional data that become increasingly common in psychological and educational research. 

This study proposes an innovative algorithm combining importance-weighted autoencoder \citep{burda2015importance} with AVB to enhance the estimation of parameters for more general and complex IFA. The proposed algorithm can be regarded as an improvement of the Importance-weighted Autoencoder (IWAE) algorithm by \citet{urban2021deep}. The main contributions are two-fold: (1) The novel Importance-weighted Adversarial Variational Bayes (IWAVB) method is proposed for exploratory and confirmatory item factor analysis of polytomous item response data; (2) Compared with IWAE method, which is the current state-of-the-art and was shown to be more efficient than the classical Metropolis-Hastings Robbins–Monro (MH-RM; \citealp{cai2010high}), numerical experiment results support similarly satisfactory performance of IWAVB, with improvement when the latent variables have a multimodal distribution.

The remaining parts are organized as follows. Section 2 explains how AVB builds upon traditional VAE methods for parameter estimation for the Graded Response Model, a general framework within Item Factor Analysis. Section 3 details the implementation steps of the AVB algorithm and introduces strategies for improving its overall performance. Section 4 then provides a comparative evaluation of AVB and VAE, using both empirical and simulated data. Finally, Section 5 examines potential limitations of the proposed approach and highlights directions for future development.

\section{Adversarial Variational Bayes for Item Factor Analysis}

In the context of IFA, given the observed responses $\bm{x}$, the Variational Inference (VI) method uses an inference model $q_{\bm{\phi}}(\bm{z}\mid\bm{x})$ to approximate the posterior distribution of the latent variable $p_{\bm{\theta}}(\bm{z}\mid\bm{x})$. The $\bm{\phi}$ represents the parameters for the inference model, and ${\bm{\theta}}$ includes the parameters for the posterior distributions (e.g., items' parameters). As a specific approach to perform VI, Adversarial Variational Bayes can be regraded as an improved version of Variational Autoencoder which constructs more expressive inference model $q_{\bm{\phi}}(\bm{z}\mid\bm{x})$, and is shown to improve the likelihood estimation of parameters for the Graded Response Model in this section.

\subsection{The Graded Response Model}
The Graded Response Model (GRM; \citealp{samejima1969estimation}) is a specific model for IFA used for analyzing ordinal response data, particularly in psychological and educational measurement. The GRM assumes responses to an item are ordered categories (e.g., Likert scale). We follow the notation as in \citet{cai2010high}.

Given $i=1,\dots,N$ distinct respondents and $j=1,\dots,M$ items, the response for respondent $i$ to item $j$ in $C_j$ graded categories is $x_{i,j}\in\left\{0,1,\dots,C_j-1\right\}$. Assuming the number of latent variables is $P$, the $P \times 1$ vector of latent variables for respondent $i$ is denoted as $\bm{z}_i$. For item $j$, the $P \times 1$ vector of loadings is $\bm{\beta}_j$, and the $(C_j-1)\times1$ vector of strictly ordered category intercepts is $\bm{\alpha}_j=(\alpha_{j,1},\dots,\alpha_{j,C_j-1})^T$. Also, the parameters of item $j$ are denoted as $\bm{\theta}_j=(\bm{\alpha}_j,\bm{\beta}_j)$. The Graded Response Model is defined as the boundary probability conditional on the item parameters $\bm{\theta}_j$ and latent variables $\bm{z}_i$: for $k\in\left\{1,\dots,C_j-1\right\}$,
\begin{equation}\label{eq:grm}
    \begin{aligned}
        p(x_{i,j}\geq k \mid \bm{z}_i,\bm{\theta}_j)&=\frac{1}{1+\exp\left[-(\bm{\beta}_j^T\bm{z}_i+\alpha_{j,k})\right]},\\
         p(x_{i,j}\geq 0 \mid \bm{z}_i,\bm{\theta}_j)&=1,\;
         p(x_{i,j}\geq C_j \mid \bm{z}_i,\bm{\theta}_j)=0.
    \end{aligned}
\end{equation}
The conditional probability for a specific response $x_{i,j}=k\in\left\{0,\dots,C_j-1\right\}$ is
\begin{equation}\label{eq:grm_prob}
    p_{i,j,k} = p(x_{i,j}= k \mid \bm{z}_i,\bm{\theta}_j)=p(x_{i,j}\geq k \mid \bm{z}_i,\bm{\theta}_j)-p(x_{i,j}\geq k+1 \mid \bm{z}_i,\bm{\theta}_j).
\end{equation}

\subsection{Marginal Maximum Likelihood}
Marginal Maximum Likelihood estimates model parameters by maximizing the likelihood of the observed response data while integrating over latent variables
. As a classical approach for parameter estimation in item response models, MML typically employs the Expectation-Maximization (EM) algorithm \citep[e.g.,][]{dempster1977maximum,bock1981marginal, muraki1992generalized, rizopoulos2007ltm, johnson2007marginal} or quadrature-based methods \citep[e.g.,][]{darrell1970fitting,rabe2005maximum}. The basic concepts of MML estimation are outlined as follows.

For the respondent $i$, and the item $j$, with the \textbf{Equation (\ref{eq:grm_prob})}, the response $x_{i,j}$ follows the multinomial distribution with 1 trail and $C_j$ mutually exclusive outcomes. The probability mass function is: $p_{\bm{\theta_j}}(x_{i,j} \mid \bm{z}_i)=\prod_{k=0}^{C_j-1}p_{i,j,k}^{\mathbbm{1}_k(x_{i,j})}$, where $\mathbbm{1}_k(x_{i,j})=1$ if $x_{i,j}=k$ and otherwise it is 0.

Considering the response pattern $\bm{x}_i=(x_{i,j})_{j=1}^M$ of the respondent $i$, since each response is conditionally independent given the latent variables, the conditional probability is: for $\bm{\theta}=(\bm{\theta}_j)_{j=1}^M$,
\begin{equation}\label{eq:pmf_xi}
    p_{\bm{\theta}}(\bm{x}_i \mid \bm{z}_i)=\prod\limits_{j=1}^M p_{\bm{\theta_j}}(x_{i,j} \mid \bm{z}_i).
\end{equation}
Given the prior distribution of latent variables is $p(\bm{z}_i)$ and $p_{\bm{\theta}}(\bm{x}_i \mid \bm{z}_i)$, the marginal likelihood of $\bm{x}_i$ is:
\begin{equation}\label{eq:mpdf_zi}
    p_{\bm{\theta}}(\bm{x}_i)=\int_{\mathbb{R}^P}\prod\limits_{j=1}^M p_{\bm{\theta_j}}(x_{i,j} \mid \bm{z}_i)p(\bm{z}_i)d\bm{z}_i.
\end{equation}
Given the full $N\times M$ responses matrix $\bm{X}=[x_{i,j}]_{i=1,\dots,N,\,j=1\dots,M}$ of all the respondents, the MML estimator $\bm{\theta}^*$ of item parameter is achieved by maximizing the likelihood of the observed responses:
\begin{equation}\label{eq:ml_X}
    \mathcal{L}(\bm{\theta}\mid\bm{X})=\prod_{i=1}^N p_{\bm{\theta}}(\bm{x}_i)=\prod_{i=1}^N\left[\int_{\mathbb{R}^P}\prod\limits_{j=1}^M p_{\bm{\theta_j}}(x_{i,j} \mid \bm{z}_i)p(\bm{z}_i)d\bm{z}_i\right].
\end{equation}
Since the above equation contains $N$ integrals in high-dimensional latent space $\mathbb{R}^P$, directly maximizing the likelihood is computationally intensive. Instead, variational methods that the current study focuses on improving, approximate $\log \mathcal{L}(\bm{\theta}\mid\bm{X})$ by one lower bound and achieve efficient computation. {This lower bound is derived based on the latent variable posterior $p_{\bm{\theta}}(\bm{z}\mid\bm{x}) = p_{\bm{\theta}}(\bm{x}\mid\bm{z})p(\bm{z})/p_{\bm{\theta}}(\bm{x})$.}

\subsection{Variational Inference and Variational Autoencoder}\label{sec:vae}
Variational methods approximate the intractable posterior $p_{\bm{\theta}}(\bm{z}\mid\bm{x})$ with a simpler inference model $q_{\bm{\phi}}(\bm{z}\mid\bm{x})$, and have gained increasing attention in the context of IFA \citep[e.g.,][]{cho2021gaussian, jeon2017variational, urban2021deep}. For distributions $p$ and $q$, Variational Inference aims to minimize the Kullback-Leibler (KL) divergence, ${\text{KL}}\left(q , p\right)$\footnote{The KL divergence is defined as ${\text{KL}}\left(q , p\right) = \mathbb{E}_q\left [\log q -\log p \right] = \int q_{\bm{\phi}}(\bm{z}\mid\bm{x}) \left [\log q_{\bm{\phi}}(\bm{z}\mid\bm{x})- \log p_{\bm{\theta}}(\bm{z}\mid\bm{x}) \right] d\bm{z} \geq 0,$ with equality if and only if $p=q$ almost everywhere with respect to $q$.}, which measures the difference between $p$ and $q$. This is related to the marginal log-likelihood of each response pattern $\bm{x}$, due to following decomposition:
\begin{equation}\label{eq:ml-decom}
    \log p_{\bm{\theta}}(\bm{x}) = \text{KL}\left[q_{\bm{\phi}}(\bm{z}\mid\bm{x}) , p_{\bm{\theta}}(\bm{z}\mid\bm{x})\right]  + \underbrace{\mathbb{E}_{q_{\bm{\phi}}(\bm{z}\mid\bm{x})}\left[ \log p_{\bm{\theta}}(\bm{z}, \bm{x}) - \log q_{\bm{\phi}}(\bm{z}\mid\bm{x})\right]}_{\text{ELBO}(\bm{x})}
\end{equation}
$\text{KL}(q,p)$ is non-negative, so the expectation term is the evidence lower bound (ELBO). For fixed $\bm{\theta}$, maximizing ELBO in $\bm{\phi}$ space is equivalent to minimizing $\text{KL}(q,p)$, while for fixed $\bm{\phi}$, increasing ELBO with respect to $\bm{\theta}$ will push the marginal likelihood $p_{\bm{\theta}}(\bm{x})$ higher. The best case is that if $q_{\bm{\phi}}(\bm{z}\mid\bm{x})$ exactly matches $p_{\bm{\theta}}(\bm{z}\mid\bm{x})$, $\text{KL}(q,p)=0$ and the MML estimator $\bm{\theta}^*$ will also be the maximizer of ELBO. The ELBO can be further decomposed:
\begin{equation}\label{eq:elbo}
    \text{ELBO}(\bm{x})=\mathbb{E}_{q_{\bm{\phi}}(\bm{z}\mid\bm{x})}\left[ \log p_{\bm{\theta}}(\bm{x} \mid \bm{z}) \right] - \text{KL}\left[ q_{\bm{\phi}}(\bm{z}\mid\bm{x}) , p(\bm{z})\right]
\end{equation}
{The first term is the negative reconstruction loss, quantifying how accurately the estimated GRM can get the original input response $\bm{x}$ from the respondent's latent $\bm{z}$. The second term is the KL divergence between $q_{\bm{\phi}}(\bm{z}\mid\bm{x})$ and $p(\bm{z})$, regularizing them to stay close.} Considering {the empirical distribution of the observed item response patterns, $p_{\bm{D}}(\bm{x})$}, our objective, $max_{\bm{\theta},\bm{\phi}} \;\mathbb{E}_{p_{\bm{D}}(\bm{x})}\left[\text{ELBO}(\bm{x})\right]$, can be further derived as: 
{
\begin{equation}\label{eq: obj1}
    \max_{\bm{\theta}} \max_{\bm{\phi}} \;\mathbb{E}_{p_{\bm{D}}(\bm{x})} 
    \mathbb{E}_{q_{\bm{\phi}}(\bm{z} \mid \bm{x})} \left( 
    \log p(\bm{z}) 
    - \log q_{\bm{\phi}}(\bm{z} \mid \bm{x}) 
    + \log p_{\bm{\theta}}(\bm{x} \mid \bm{z}) 
    \right).
\end{equation}
}

Variational Autoencoder (VAE; \citealp{kingma2013auto,urban2021deep}) leverages VI to learn a probabilistic representation of observed variables and generate new samples. It combines ideas from deep learning via neural networks and statistical inference. The latent variables $\bm{z}$ are typically assumed to follow a standard normal distribution $\mathcal{N}(\bm{0},\bm{I})$, where $\bm{I}$ is the identity matrix. The simpler and tractable distribution $q_{\bm{\phi}}(\bm{z}\mid\bm{x})$ for continuous latent variables $\bm{z}$ is parameterized by a neural network as the encoder. A typical choice is the isotropic normal distribution:
\begin{equation}\label{eq:iso-normal}
    q_{\bm{\phi}}(\bm{z}\mid\bm{x})=q_{\bm{\phi}(\bm{x})}(\bm{z})=\mathcal{N}(\bm{z} \mid \bm{\mu}(\bm{x}),\bm{\sigma}^2(\bm{x})\bm{I}),
\end{equation}
where $\bm{\phi}(\bm{x})=(\bm{\mu}(\bm{x}),\bm{\sigma}(\bm{x}))$ denotes the normal distribution's configuration in the form of vector function of $\bm{x}$ for convenience; $\bm{x}$, $\bm{\mu}(\bm{x})$, and $\bm{\sigma}(\bm{x})$ are vectors with the same dimension. Given the latent representation $\bm{z}$, the decoder is the probabilistic model which reconstructs or generates data via $p_{\bm{\theta}}(\bm{x} \mid \bm{z})$. However, the inference model expressed in the \textbf{Equation (\ref{eq:iso-normal})} might not be expressive enough to capture the true posterior \citep{mescheder2017adversarial}. The KL divergence term in the ELBO requires explicit forms for both $q_{\bm{\phi}}(\bm{z}\mid\bm{x})$ and $p(\bm{z})$, which can limit the choice of posterior families. Highly expressive inference models can lead to higher log-likelihood bounds and stronger decoders making full use of the latent space \citep{kingma2016improved,chen2016variational}. {One remedy is to replace the inference model with an implicit distribution that lacks a closed-form density and to approximate the intractable $\text{KL}\left[ q_{\bm{\phi}}(\bm{z}\mid\bm{x}) , p(\bm{z})\right]$ via auxiliary networks \citep{shi2018kernel}. AVB is a representative example of this approach.}

\subsection{Adversarial Variational Bayes}\label{sec:avb}
\begin{figure}[htb]
\centering
\includegraphics[width=0.95\textwidth]{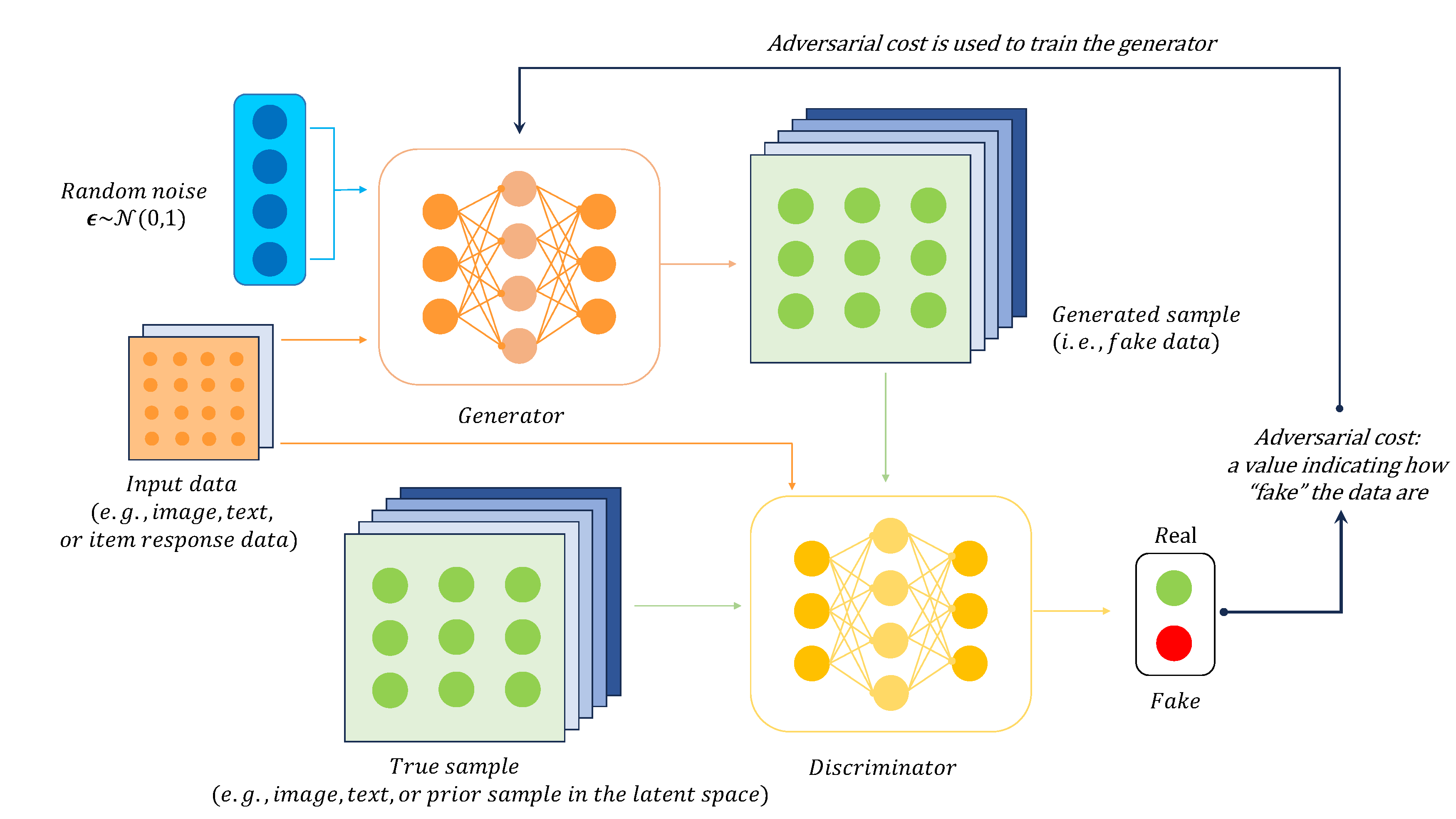}
\caption{Schematic illustration of a standard Generative Adversarial Network. {In some GAN variants, real data serve only as true samples and are not fed into the generator. However, in the AVB application to IFA, the generator and discriminator take item response data as input, and the discriminator distinguishes between samples in the latent space.}}
\label{fig:gan}
\end{figure}

Adversarial Variational Bayes borrows ideas from the Generative Adversarial Network framework, which leads to remarkable results in areas such as image generation \citep{goodfellow2014generative} and data augmentation \citep{zhu2017data}. As shown in the \textbf{Figure \ref{fig:gan}}, GANs employ adversarial training by pitting a generator against a discriminator, compelling each to enhance its performance. This competition results in highly convincing synthetic data, highlighting GANs' power in discovering intricate patterns found in real-world samples. {In the context of IFA, item response data are inputs to both the generator and discriminator. Random noise sampled from the prior distribution is also required for the generator to get ``generated'' latent variables, which, along with ``true'' samples drawn directly from the prior distribution, are then evaluated by the discriminator.}

To integrate GANs into VAE, AVB regards the inference model $q_{\bm{\phi}}(\bm{z}\mid\bm{x})$ as the generator of latent variables and replaces the explicit KL divergence computation in the VAE with an adversarially trained discriminator, allowing for more expressive inference models. To be specific, besides the \textbf{Encoder} network $q_{\bm{\phi}}(\bm{z}\mid\bm{x})$ and the \textbf{Decoder} network $p_{\bm{\theta}}(\bm{x} \mid \bm{z})$, a neural network {$T_{\bm{\psi}}(\bm{x}, \bm{z})$ with parameters $\bm{\psi}$} as the \textbf{Discriminator} learns to distinguish between samples from the inference model $q_{\bm{\phi}}(\bm{z} \mid \bm{x})$ and the prior $p(\bm{z})$ {via minimizing the binary classification loss, which can be formulated as the following objective:}
\begin{equation}\label{eq:T}
\max_{\bm{\psi}} \, \mathbb{E}_{p_{\bm{D}}(\bm{x})} 
\mathbb{E}_{q_{\bm{\phi}}(\bm{z} \mid \bm{x})} 
\log \sigma \left( T_{\bm{\psi}}(\bm{x}, \bm{z}) \right) 
+ \mathbb{E}_{p_{\bm{D}}(\bm{x})} 
\mathbb{E}_{p(\bm{z})} 
\log \left( 1 - \sigma \left( T_{\bm{\psi}}(\bm{x}, \bm{z}) \right) \right). 
\end{equation}
Here, $\sigma(t) := \left(1 + e^{-t}\right)^{-1}$ denotes the sigmoid-function. Intuitively, $T_{\bm{\psi}}(\bm{x}, \bm{z})$ tries to distinguish pairs $(\bm{x}, \bm{z})$ that are sampled independently using the distribution $p_{\bm{D}}(\bm{x}) p(\bm{z})$ from those that are sampled using the current inference model (i.e., $p_{\bm{D}}(\bm{x}) q_{\bm{\phi}}(\bm{z} \mid \bm{x})$). The \textbf{Equation (\ref{eq:T})} encourages samples from $q_{\bm{\phi}}(\bm{z} \mid \bm{x})$ but punishes samples from $p(\bm{z})$. {As shown by \citet{mescheder2017adversarial}, the KL term in the ELBO can be recovered by the following theorem.}
\begin{theorem}[{\citealp{mescheder2017adversarial}}]\label{thm:T_1}
    For $p_{\bm{\theta}}(\bm{x} \mid \bm{z})$ and $q_{\bm{\phi}}(\bm{z} \mid \bm{x})$ fixed, the optimal discriminator $T^*$ according to the \textbf{Objective (\ref{eq:T})} is given by
    \begin{equation}\label{eq:T_thm}
    T^*(\bm{x}, \bm{z}) = \log q_{\bm{\phi}}(\bm{z} \mid \bm{x}) - \log p(\bm{z}), 
    \end{equation}
    so $\text{KL}\left[ q_{\bm{\phi}}(\bm{z}\mid\bm{x}), p(\bm{z})\right] = \mathbb{E}_{q_{\bm{\phi}}(\bm{z} \mid \bm{x})} \left(T^*(\bm{x}, \bm{z})\right)$.
\end{theorem}
On the other hand, the encoder $q_{\bm{\phi}}(\bm{z} \mid \bm{x})$ is trained adversarially to ``fool'' the discriminator, since if $q_{\bm{\phi}}(\bm{z} \mid \bm{x})$ is indistinguishable from $p(\bm{z})$, then they can be regarded as similar distributions. Using the reparameterization trick, the \textbf{Objective (\ref{eq: obj1})} can be rewritten in the form:
{
\begin{equation}\label{eq: obj2}
    \max_{\bm{\theta}} \max_{\bm{\phi}} \; \mathbb{E}_{p_{\bm{D}}(\bm{x})} \mathbb{E}_{\bm{\epsilon}} 
    \left( - {T}^*\left( \bm{x}, \bm{z}_{\bm{\phi}}(\bm{x}, \bm{\epsilon}) \right) 
    + \log p_{\bm{\theta}} \left( \bm{x} \mid \bm{z}_{\bm{\phi}}(\bm{x}, \bm{\epsilon}) \right) 
    \right)
\end{equation}
}
Therefore, we need to do two optimizations for \textbf{Objectives (\ref{eq:T})} and \textbf{(\ref{eq: obj2})}, summarized in \textbf{Algorithm \ref{alg:avb}}. Note that \( T_{\bm{\psi}}(\bm{x}, \bm{z}) \) and \( \bm{z}_{\bm{\phi}}(\bm{x}, \bm{\epsilon}) \) can be implemented as neural networks, but in the context of the GRM, the decoder \( p_{\bm{\theta}} \left( \bm{x} \mid \bm{z}_{\bm{\phi}}(\bm{x}, \bm{\epsilon}) \right) \) corresponds directly to the likelihood definition for the GRM. As such, it is modeled by a single-layer neural network consisting of a linear function and a sigmoid activation. 
\begin{algorithm}[!htb]
\caption{Adversarial Variational Bayes (AVB)}
\SetAlgoLined

\SetKwInOut{Given}{Given}
\Given{Responses $\bm{x}_1,\dots,\bm{x}_N$}
\While{not converged}{
    Sample $\bm{z}_1,\dots,\bm{z}_N$ from prior $p(\bm{z})$\;
    Sample $\bm{\epsilon}_1,\dots,\bm{\epsilon}_N$ from $\mathcal{N}(0,1)$\;
    Compute $\bm{\theta}$-gradient (\textbf{Objective (\ref{eq: obj2})}):
    $g_{\bm{\theta}} \leftarrow \frac{1}{N}\sum_{i=1}^N\nabla_{\bm{\theta}} \log p_{\bm{\theta}} \left( \bm{x}_i \mid \bm{z}_{\bm{\phi}}(\bm{x}_i, \bm{\epsilon}_i) \right) $\;
    Compute $\bm{\phi}$-gradient (\textbf{Objective (\ref{eq: obj2})}):
    $g_{\bm{\phi}} \leftarrow \frac{1}{N}\sum_{i=1}^N\nabla_{\bm{\phi}} \left[ - {T}_{\bm{\psi}}\left( \bm{x}_i, \bm{z}_{\bm{\phi}}(\bm{x}_i, \bm{\epsilon}_i) \right) + \log p_{\bm{\theta}} \left( \bm{x}_i \mid \bm{z}_{\bm{\phi}}(\bm{x}_i, \bm{\epsilon}_i) \right) \right]  $\;
    Compute $\bm{\psi}$-gradient (\textbf{Objective (\ref{eq:T})}): 
    $g_{\bm{\psi}} \leftarrow \frac{1}{N}\sum_{i=1}^N\nabla_{\bm{\psi}} \left[ \log \sigma \left( T_{\bm{\psi}}(\bm{x}_i, \bm{z}_{\bm{\phi}}(\bm{x}_i, \bm{\epsilon}_i)) \right) + \log \left( 1 - \sigma \left( T_{\bm{\psi}}(\bm{x}_i, \bm{z}_i) \right) \right) \right]  $\;
    Perform SGD-updates for $\bm{\theta}$, $\bm{\phi}$ and $\bm{\psi}$\;
}
\label{alg:avb}
\end{algorithm}

\subsection{Our Contributions to the Literature}
Given that VAEs have been a powerful approach to perform exploratory IFA \citep{urban2021deep}, the innovative contribution of the AVB method is to bring richer inference model choices for the implementation by deep learning. Over the past decade, deep learning techniques have thrived \citep{lecun2015deep}, and IFA has derived considerable benefits from these advances. When applying neural networks to estimate parameters in IFA, key distinctions arise in the design of the network architecture and the choice of loss function. A straightforward strategy is to feed one-hot encodings\footnote{In one‐hot encoding, each respondent or item is represented by a binary vector whose length matches the total number of respondents or items. Exactly one element of this vector is set to 1 (indicating the specific respondent or item), and all other elements are set to 0.} for both respondents and items into two separate feedforward neural networks: one infers the latent variables of the respondents, while the other estimates the item parameters. This approach typically employs a variation of the Joint Maximum Likelihood (JML) for its loss function, relying on the product of conditional probabilities in \textbf{Equation (\ref{eq:pmf_xi})} across all respondents \citep{tsutsumi2021deep,zhang2024neural}. However, as demonstrated by \citet{ghosh1995inconsistent}, the JML solution leads to inconsistent estimates for item parameters.

An alternative that proves more statistically robust relies on the marginal maximum likelihood procedure discussed earlier \citep{bock1981marginal,wirth2007item}. Drawing on these perspectives, researchers merge representation learning for latent-variable models with neural architectures by adopting autoencoder designs. Through variational methods, they approximate maximum likelihood estimation for latent variable models. A notable example in the literature is the Importance-Weighted Autoencoder algorithm \citep{urban2021deep}. In our study, we propose an importance-weighted adversarial variational Bayes, refines this approach.

\begin{figure}[!htb]
    \centering
    \subfigure[Variational Autoencoder]{
        \includegraphics[width=0.9\textwidth]{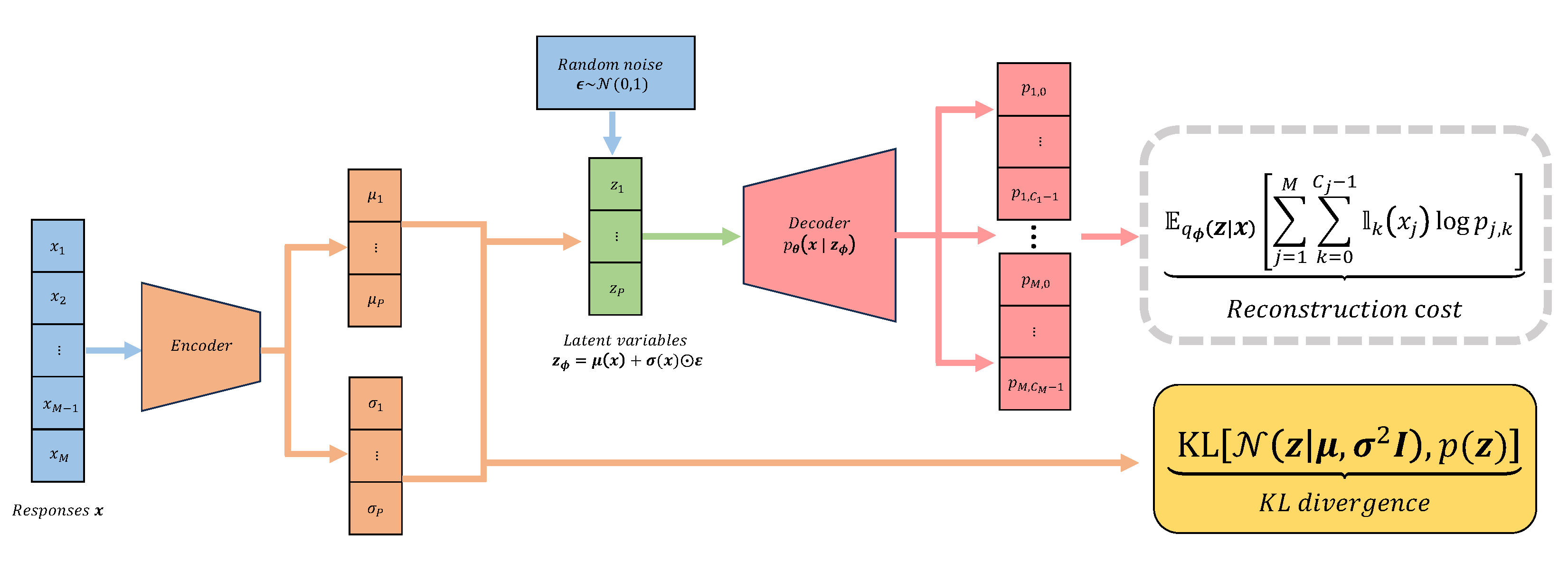}
        \label{fig:comp_vae}}

    \subfigure[Adversarial Variational Bayes]{
        \includegraphics[width=0.9\textwidth]{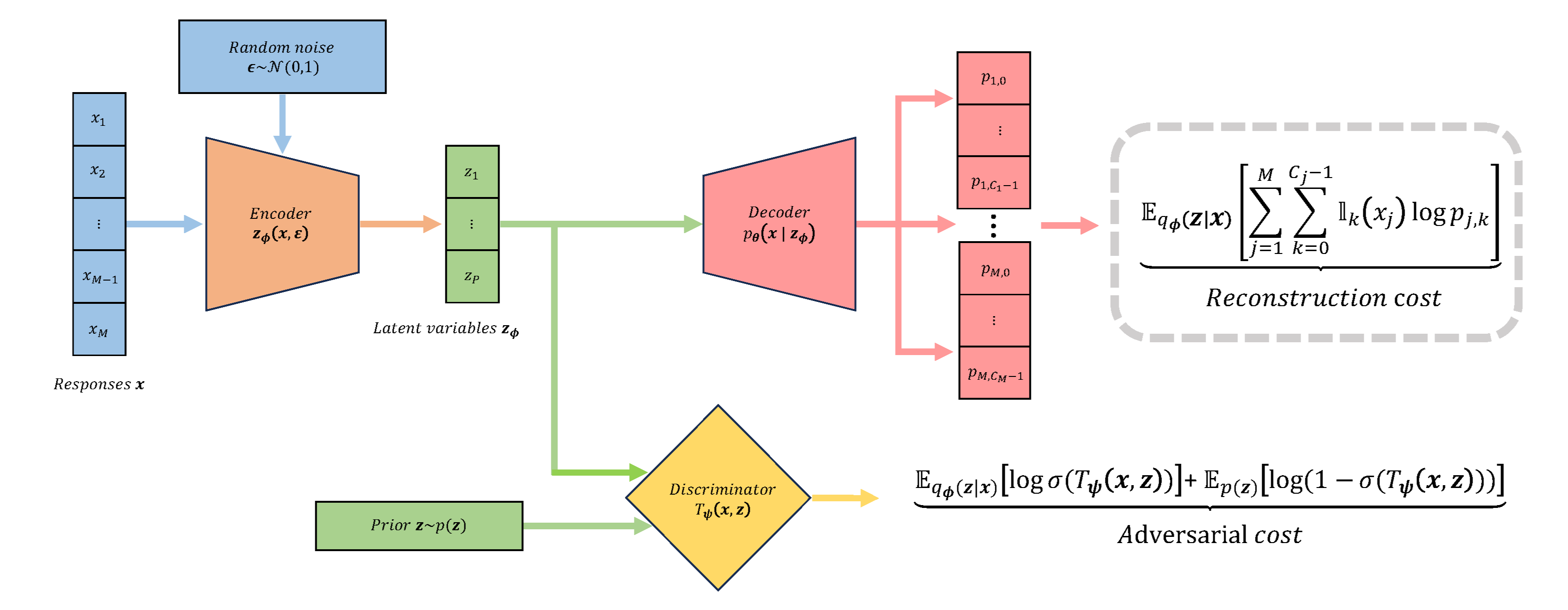}
        \label{fig:comp_avb}}

    \caption{Schematic comparison of the encoder and decoder designs for the AVB method and a standard VAE.}
    \label{fig:comp_avb_vae}
\end{figure}

Building on insights from Sections \ref{sec:vae} and \ref{sec:avb}, we observe that the major contrast between AVB and conventional VAE approaches centers on the inference model. {\textbf{Figure \ref{fig:comp_avb_vae}} visually contrasts these two approaches.} While generative models demand precise techniques for latent-variable estimation, basic Gaussian assumptions might restrict the range of distributions in standard VAEs. Although researchers have attempted to design more sophisticated neural network architectures for VAE inference, these solutions may still lack the expressiveness of the black-box method employed by AVB \citep{mescheder2017adversarial}. 

{To address this limitation, the AVB framework modifies $q_{\bm{\phi}}(\bm{z}\mid\bm{x})$ into a fully black-box neural network $\bm{z}_{\bm{\phi}}(\bm{x}, \bm{\epsilon})$. Rather than a standard approach where random noise is added only at the final step, AVB incorporates $\bm{\epsilon}$ as an additional input earlier in the inference process. As neural networks can almost represent any probability density in the latent space \citep{cybenko1989approximation}, this design empowers the network to represent intricate probability distributions free from the Gaussian assumption, thus expanding the scope of patterns it can capture. However, such implicit $q_{\bm{\phi}}$ lacks a tractable density, $\text{KL}\left[ q_{\bm{\phi}}(\bm{z}\mid\bm{x}) , p(\bm{z})\right]$ cannot be directly computed so another discriminator neural network learns to recover $\text{KL}\left[ q_{\bm{\phi}}(\bm{z}\mid\bm{x}) , p(\bm{z})\right]$ via adversarial process.} {The recovery of $\text{KL}\left[ q_{\bm{\phi}}(\bm{z}\mid\bm{x}) , p(\bm{z})\right]$ imposes a prior, such as a standard normal, on $\bm{z}$. This prior anchors the location and scale of $\bm{z}$, thereby alleviating the translational and scaling indeterminacies that cause model non-identifiability \citep{de2013theory}.}

Through the adversarial process, the algorithm derives item parameter estimates $\bm{\theta}^*$ that optimizes the likelihood of the observed data, while also constructing a distribution $q_{\bm{\phi}^*}(\bm{z}\mid\bm{x})$ that closely approximates the true posterior $p_{\bm{\theta}^*}(\bm{z}\mid\bm{x})$. From a theoretical perspective, \citet{mescheder2017adversarial} established that AVB can achieve performance \textit{\textbf{on par with at least that of VAEs}} in terms of likelihood, as evidenced by the following theorem:
\begin{theorem}[{\citealp{mescheder2017adversarial}}]\label{thm:T_2}
    Assume that $T$ can represent any function of $\bm{x}$ and $\bm{z}$. If $(\bm{\theta}^*,\bm{\phi}^*,T^*)$ defines a Nash-equilibrium\footnote{Nash equilibrium is a central concept in game theory wherein each player’s strategy is optimal, given the strategies of all other players. In the context of GANs, this equilibrium arises in the two‐player adversarial optimization: once the generator and the discriminator reach a Nash equilibrium, neither can unilaterally adjust its parameters to further improve its objective without a corresponding change by the other.} of the two-player game defined by the \textbf{Equations (\ref{eq:T}) and (\ref{eq: obj2})}, then
    \begin{equation}\label{eq:opt_T}
        T^*(\bm{x},\bm{z})=\log q_{\bm{\phi}^*}(\bm{z}\mid\bm{x}) - \log p(\bm{z})
    \end{equation}
    and $(\bm{\theta}^*,\bm{\phi}^*)$ is a global optimum of the variational lower bound in the \textbf{Equation (\ref{eq: obj1})}.
\end{theorem}

Under misspecification such as inappropriate assumptions of latent variable distributions, AVB’s architecture can deliver a higher marginal likelihood theoretically. Meanwhile, by improving the marginal log-likelihood, one can reduce the reconstruction error $-\mathbb{E}_{q_{\bm{\phi}}(\bm{z}\mid\bm{x})}\left[ \log p_{\bm{\theta}}(\bm{x} \mid \bm{z}) \right]$, which measures the disparity between real and reconstructed data, and can be reduced by increasing marginal likelihood according to \textbf{Equation (\ref{eq:elbo})}. In the context of IRT, AVB thus stands out by capturing a more accurate respondents’ latent traits. Research in the computer science field also highlights that the combination of GAN and VAE can extract more discriminative information and provide more precise and clear model in the latent variable space \citep[e.g.,][]{makhzani2015adversarial,hou2019improving}.

\section{Implementation of AVB and IWAVB}
This section introduces the implementation details of the AVB method. A fundamental neural network design is provided first and to achieve better approximation and higher likelihood, the adaptive contrast and importance-weighted techniques are introduced. A summary with a detailed algorithm combining all the techniques is shown in the end.

\subsection{Neural Network Design}
Considering the schematic diagram in the \textbf{Figure \ref{fig:comp_avb}}, the encoder $\bm{z}_{\bm{\phi}}(\bm{x}, \bm{\epsilon})$ and discriminator $T_{\bm{\psi}}(\bm{x}, \bm{z})$ are designed to be multiple-layer feedforward neural networks (FNNs) in this study. A FNN is a type of artificial neural network where information flows in one direction, from the input layer to the output layer, through one or more hidden layers, without forming any cycles or loops. It is the simplest form of a neural network and by increasing the number of neurons in the hidden layer, any continuous function on a closed interval can be approximated at an arbitrary precision \citep{cybenko1989approximation}.

FNN consists of layers of neurons, where each layer 
$l$ is represented mathematically by the following recursive relation: given the weight matrix $\bm{W}^{(l)}$ connecting layer $l-1$ and $l$, the $P_{l-1}\times1$ output vector $\bm{h}^{(l-1)}$ from the previous layer $l-1$ and the $P_l\times1$ bias vector $\bm{b}^{(l)}$ for layer $l$, the $P_{l}\times1$ output $\bm{h}^{(l)}$ of the neurons in layer $l$ is defined by
\begin{equation}\label{eq:fnn}
    \bm{h}^{(l)} = f^{(l)}(\bm{W}^{(l)}\bm{h}^{(l-1)}+\bm{b}^{(l)}).
\end{equation}
For $l=0$, $\bm{h}^{(0)}=\bm{x}$, the input vector to the network. If the total number of layers is $L$, $\bm{h}^{(1)},\dots,\bm{h}^{(L-1)}$ are the hidden layers and the final output vector of the FNN is $\bm{h}^{(L)}$. Therefore, the $\bm{\phi}$ for encoder and $\bm{\psi}$ for discriminator include a set of weight matrices $\bm{W}^{(l)}$ and bias vectors $\bm{b}^{(l)}$ for $l=1,\dots,L$ respectively.

The activation function $f(\cdot)$ applied element-wise introduces non-linearity into the network, enabling it to approximate complex functions. In this research, the Gaussian Error Linear Unit (GELU; \citealp{hendrycks2016gaussian}) is set as the activation functions $f^{(1)},\dots,f^{(L-1)}$ for the hidden layers in the encoder network and discriminator network:
\begin{equation}\label{eq:glu}
    \text{GELU}(x) = x \cdot \Phi(x),
\end{equation}
where $\Phi(x)$ is the cumulative distribution function of the standard normal distribution. GELU is continuously differentiable, which facilitates stable gradient propagation during training. Studies have demonstrated that models utilizing GELU outperform those with ReLU and Exponential Linear Unit (ELU) activations across various tasks, including computer vision and natural language processing \citep{hendrycks2016gaussian}. The final activation function for the output of the network is set to the identity function $f^{(L)}(x)=x$.

{
Parameter initialization is essential for training neural networks, as it sets initial values for weights and biases to ensure effective learning. Proper initialization prevents common training issues, such as vanishing or exploding gradients. In this study, Kaiming initialization \citep{he2015delving} was chosen for the encoder and discriminator networks, as it is specifically designed for asymmetric activations such as GELU, maintaining stable variance across layers. For the decoder, viewed as a single-layer neural network with a sigmoid activation, Xavier initialization \citep{glorot2010understanding} was applied to loading and intercept parameters to balance variance effectively.
}

{
Neural network parameters are commonly optimized using Stochastic Gradient Descent (SGD), an iterative method that updates parameters by minimizing a loss function computed from random batches of training data \citep{lecun2015deep}. Formally, parameters $\bm{\omega}_t$ (e.g., $\bm{\theta},\bm{\phi},\bm{\psi}$ in the \textbf{Algorithm \ref{alg:avb}}) are updated according to $\bm{\omega}_{t+1} = \bm{\omega}_t - \eta \nabla_{\bm{\omega}} \mathcal{L}(\bm{\omega})$, where $\eta$ is the learning rate, and $\nabla_{\bm{\omega}} \mathcal{L}(\bm{\omega})$ is the gradient. Modern extensions of SGD methods, such as AdamW \citep{loshchilov2017decoupled}, improve on SGD by adaptive learning rates and decoupled weight decay. AdamW has become a popular choice in optimizing neural networks due to its ability to converge faster and improve model performance by effectively mitigating overfitting caused by large parameter values \citep{zhuang2022understanding}. Further details on initialization and AdamW optimization are provided in \nameref{sec:appendix}.
}

In the \textbf{Algorithm \ref{alg:avb}}, the gradient descent step tries to force $T(\bm{x},\bm{z})$ close to $T^*(\bm{x},\bm{z})$ but the gap might fail to be sufficiently small. {$T(\bm{x},\bm{z})$ tries to recover $\text{KL}\left[ q_{\bm{\phi}}(\bm{z}\mid\bm{x}) , p(\bm{z})\right]$, but this task can be quite challenging if there’s a large difference between $q_{\bm{\phi}}(\bm{z}\mid\bm{x})$ and the prior distribution $p(\bm{z})$. Therefore, \citet{mescheder2017adversarial} also provided a technique to replace $\text{KL}\left[ q_{\bm{\phi}}(\bm{z}\mid\bm{x}) , p(\bm{z})\right]$ with another KL divergence between $q_{\bm{\phi}}(\bm{z}\mid\bm{x})$ and an adaptive distribution $r_{\alpha}(\bm{z}\mid\bm{x})$ rather than the simple, fixed Gaussian $p(\bm{z})$.}

\subsection{Adaptive Contrast Technique}
Adaptive Contrast (AC; \citealp{mescheder2017adversarial}) is a technique to handle the issue of large gap between $q_{\bm{\phi}}(\bm{z}\mid\bm{x})$ and $p(\bm{z})$. Instead of contrasting $q_{\phi}(\bm{z}\mid\bm{x})$ with $p(\bm{z})$, an auxiliary conditional distribution $r_{\alpha}(\bm{z}\mid\bm{x})$ serves as an intermediate step to bridge their gap. Then the \textbf{Objective (\ref{eq: obj1})} can be rewritten as:
\begin{equation}\label{eq: obj-AC}
\max_{\bm{\theta}} \max_{\bm{\phi}} \;\mathbb{E}_{p_{\bm{D}}(\bm{x})} \left[ 
- \text{KL}\left[ q_{\bm{\phi}}(\bm{z}\mid\bm{x}), r_{\alpha}(\bm{z}\mid\bm{x})\right]
+ \mathbb{E}_{q_{\bm{\phi}}(\bm{z} \mid \bm{x})}\left[-\log r_{\alpha}(\bm{z}\mid\bm{x}) + \log p_{\bm{\theta}}(\bm{x} , \bm{z}) 
\right]\right].
\end{equation}
Also, the $\theta$ is updating for $p_{\bm{\theta}}(\bm{x} , \bm{z})$ instead of $p_{\bm{\theta}}(\bm{x} \mid \bm{z})$ so it will also be easier to update correlation between different dimensions of latent variables.

To closely approximate the $q_{\phi}(\bm{z}\mid\bm{x})$, a tractable density choice of $r_{\alpha}(\bm{z}\mid\bm{x})$ is a Gaussian distribution with a diagonal covariance matrix, which matches the mean $\bm{\mu}(\bm{x})$ and variance $\bm{\sigma}^2(\bm{x})$ of $q_{\phi}(\bm{z}\mid\bm{x})$. Therefore, $r_{\alpha}(\bm{z}\mid\bm{x})$ is adaptive according to updating $\phi$. The KL term can also be approximated by the \textbf{Theorem \ref{thm:T_1}} and $\text{KL}\left[ q_{\bm{\phi}}(\bm{z}\mid\bm{x}), r_{\alpha}(\bm{z}\mid\bm{x})\right]$ can be much smaller than $\text{KL}\left[ q_{\bm{\phi}}(\bm{z}\mid\bm{x}), p(\bm{z})\right]$, which makes the construction of an effective discriminator easier \citep{mescheder2017adversarial}. {Moreover, given the distribution of the normalized vector $\tilde{\bm{z}}$, $\tilde{q}_{\bm{\phi}}(\tilde{\bm{z}}\mid\bm{x})$,} reparameterization trick can further simplify the KL term into 
\begin{equation}\label{eq:AC-rep}
    \text{KL}\left[\tilde{q}_{\bm{\phi}}(\tilde{\bm{z}}\mid\bm{x}), r_{0}(\tilde{\bm{z}})\right],\text{ where } r_{0}(\tilde{\bm{z}})\sim \mathcal{N}(\bm{0},\bm{I}), \tilde{\bm{z}} = \frac{\bm{z} - \bm{\mu}(\bm{x})}{\bm{\sigma}(\bm{x})}.
\end{equation}
By using this reparameterization, we allow $T(\bm{x},\bm{z})$ to focus only on deviations of $q_{\bm{\phi}}(\bm{z}\mid\bm{x})$ from a Gaussian distribution's shape, rather than its location and scale. This refinement reduces the complexity of the adversarial task and improves the stability and accuracy of the learning process, ultimately enhancing the quality of the variational inference approximation. Moreover, as $p_{\bm{\theta}}(\bm{x},\bm{z}) = p_{\bm{\theta}}(\bm{x} \mid \bm{z})p(\bm{z})$, compared to the \textbf{Objective (\ref{eq: obj1})}, the \textbf{Objective (\ref{eq: obj-AC})} can simultaneously estimate the factor correlation $\bm{\Sigma}$ if $\bm{\Sigma}$ can be introduced into $p(\bm{z})$ as a new parameter.

\subsection{Importance-weighted Technique}
Importance-weighted Variational Inference (IWVI; \citealp{burda2015importance}) connects VI with the MML estimation. Instead of maximizing the ELBO in the \textbf{Equation (\ref{eq:ml-decom})}, the amortized IWVI now maximizes the Importance-weighted ELBO (IW-ELBO):
\begin{align}
    \bm{z}_{1:R} &\sim \prod\limits_{r=1}^R q_{\bm{\phi}}(\bm{z}_r \mid \bm{x}), w_r = p_{\bm{\theta}}(\bm{z}_r,\bm{x}) / q_{\bm{\phi}}(\bm{z}_r \mid \bm{x})\nonumber\\
    \text{IW-ELBO} &= \mathbb{E}_{\bm{z}_{1:R}}\left[\log \frac{1}{R}\sum\limits_{r=1}^R w_r\right] \leq \log p_{\bm{\theta}}(\bm{x}).\label{eq:iw-elbo}
\end{align}
If the number of importance-weighted (IW) samples $R=1$, the IW-ELBO is reduced to the ELBO. As $R$ increases, IW-ELBO becomes more similar to the marginal likelihood than ELBO \citep{burda2015importance}. Therefore, optimizing IW-ELBO, which is the importance-weighted \textbf{Objective (\ref{eq: obj1}) or (\ref{eq: obj-AC})}, achieves a better approximation to the MML estimator with slightly {lower} computational efficiency due to more samples but GPU computation source can ensure the computation time under control and even less than the time required by traditional CPU computation.

The unbiased estimator for the gradient of the IW-ELBO w.r.t $\bm{\xi}=(\bm{\theta}^T,\bm{\phi}^T)$ can be represented as follows:
\begin{equation}\label{eq:grad_iwvi}
    \nabla_{\bm{\xi}}\mathbb{E}_{\bm{z}_{1:R}}\left[ \log \frac{1}{R}\sum\limits_{r=1}^R w_r\right] = \mathbb{E}_{\bm{\epsilon}_{1:R}}\left[ \sum\limits_{r=1}^R \tilde{w}_r \nabla_{\bm{\xi}} \log w_r\right]\approx \frac{1}{S}\sum\limits_{s=1}^S\left[ \sum\limits_{r=1}^R \tilde{w}_{r,s} \nabla_{\bm{\xi}} \log w_{r,s}\right],
\end{equation}
where $\bm{\epsilon}_{1:R}\sim\prod_{r=1}^R\mathcal{N}(\bm{\epsilon}_r)$ and $\tilde{w}_r = w_r / \sum_{r'=1}^R w_{r'}$ \citep{urban2021deep}. However, increasing $R$ might make the gradient estimator with respect to $\bm{\phi}$ in the encoder network become completely random and lead to inefficient computation \citep{rainforth2018tighter}. The $\bm{\theta}$ in the decoder does not have this problem. Doubly reparameterized gradient (DReG; \citealp{tucker2018doubly,urban2021deep}) estimator can be used to change $\nabla_{\bm{\phi}}\mathbb{E}_{\bm{z}_{1:R}}\left[ \log \frac{1}{R}\sum_{r=1}^R w_r\right]$ which attains lower variance:
\begin{equation}\label{eq:dreg}
    \begin{aligned}
        \nabla_{\bm{\phi}}\mathbb{E}_{\bm{z}_{1:R}}\left[ \log \frac{1}{R}\sum\limits_{r=1}^R w_r\right] &= \mathbb{E}_{\bm{\epsilon}_{1:R}}\left[ \sum\limits_{r=1}^R \tilde{w}_r^2 \frac{\partial \log w_r}{\partial \bm{z}_r}\frac{\partial\bm{z}_r}{\partial\bm{\phi}}\right]\\
        &\approx \frac{1}{S}\sum\limits_{s=1}^S\left[ \sum\limits_{r=1}^R \tilde{w}_{r,s}^2 \frac{\partial \log w_{r,s}}{\partial \bm{z}_{r,s}}\frac{\partial\bm{z}_{r,s}}{\partial\bm{\phi}}\right].
    \end{aligned}
\end{equation}
One single Monte Carlo sample (i.e., $S=1$) is enough to attain satisfactory approximation to the gradient estimators in practice \citep{burda2015importance,tucker2018doubly,urban2021deep}.

\subsection{Implementation Summary}\label{sec:implement}
In summary, \textbf{Algorithm \ref{alg:iwavb}} shows the implementation details of the Importance-weighted Adversarial Variational Bayes method. 
\begin{algorithm}[!htb]
\caption{AVB with Adaptive Contrast and Importance-weighted techniques}
\SetAlgoLined

\SetKwInOut{Given}{Given}
\Given{Responses $\bm{x}_1,\dots,\bm{x}_N$; dimension of latent variable $P$; mini-batch size $B$; IW samples $R$; MC samples $S$; hyperparameters for AdamW $\beta_1$ and $\beta_2$; learning rate for the encoder and decoder $\eta_1$; learning rate for the discriminator $\eta_2$; initialization parameters $(\bm{\theta}_0, \bm{\phi}_0, \bm{\psi}_0)$ with Kaiming and Xavier Initialization (preventing issues like vanishing or exploding gradients); prior distribution $p(\bm{z})$ (e.g., multivariate standard Normal distribution)}
\For{iteration $t=0$ \KwTo $T$}{
    \emph{Computation step}: Randomly sample a mini-batch $\{\bm{x}_i\}_{i=1}^B$\;
    \For{respondent $i=1$ \KwTo $B$}{
        \For{IW sample and MC sample $r,s=1$ \KwTo $R,S$}{
        Sample $\bm{\epsilon}_{i,r,s} \sim \mathcal{N}(0,1)$\;
        Sample $\bm{\zeta}_{i,r,s} \sim \mathcal{N}(0,1)$\;
        $\bm{z}_{i,r,s}$, $\bm{\mu}_{i,r,s}$, $\bm{\sigma}_{i,r,s} \leftarrow \text{Encoder}_{\bm{\phi}}(\bm{x}_i,\bm{\epsilon}_{i,r,s})$\;
        $\tilde{\bm{z}}_{i,r,s} \leftarrow \frac{\bm{z}_{i,r,s}-\bm{\mu}_{i,r,s}}{\bm{\sigma}_{i,r,s}}$\;
        $w_{i,r,s} \leftarrow \exp\left[ - {T}_{\bm{\psi}}\left( \bm{x}_i, \tilde{\bm{z}}_{i,r,s} \right) + \frac{1}{2}||\tilde{\bm{z}}_{i,r,s}||^2+\log p_{\bm{\theta}} \left( \bm{x}_i, \bm{z}_{i,r,s} \right)\right]$\;
        }
        $\text{IW-ELBO}_i \leftarrow \frac{1}{S}\sum_{s=1}^S \left[\log\frac{1}{R}\sum_{r=1}^R w_{i,r,s}\right]$\;
        Objective for the discriminator: $d_i \leftarrow \frac{1}{S}\sum_{s=1}^S \left\{\frac{1}{R}\sum_{r=1}^R\left[\log \sigma \left( T_{\bm{\psi}}(\bm{x}_i, \tilde{\bm{z}}_{i,r,s} \right) + \log \left( 1 - \sigma \left( T_{\bm{\psi}}(\bm{x}_i, \bm{\zeta}_{i,r,s}) \right) \right)\right]\right\}$\;
    }
    Compute $\bm{\theta}$-gradient:
    $\bm{g}_{\bm{\theta}_t} \leftarrow \frac{1}{B}\sum_{i=1}^B\nabla_{\bm{\theta}_t} \text{IW-ELBO}_i $\;
    Compute $\bm{\phi}$-gradient:
    $\bm{g}_{\bm{\phi}_t} \leftarrow \frac{1}{B}\sum_{i=1}^B\nabla_{\bm{\phi}_t} \text{IW-ELBO}_i $\;
    Compute $\bm{\psi}$-gradient: 
    $\bm{g}_{\bm{\psi}_t} \leftarrow \frac{1}{B}\sum_{i=1}^B\nabla_{\bm{\psi}_t} d_i  $\;
    \emph{Optimization step:} Perform SGD-updates via AdamW for $\bm{g}_t = (\bm{g}_{\bm{\theta}_t}^T, \bm{g}_{\bm{\phi}_t}^T, \bm{g}_{\bm{\psi}_t}^T)$\;
}
\label{alg:iwavb}
\end{algorithm}

Determining an optimal neural network configuration (e.g., the number and size of hidden layers) and other hyper-parameters, such as learning rates, are critical for satisfactory performance. {GAN-based methods like AVB are particularly sensitive to these settings since they involve two interacting neural networks \citep{roth2017stabilizing,lucic2018gans}.} However, there is no panacea solution, as the architecture often depends on the specific application and the characteristics of the data. Empirical studies have shown that one or two hidden layers are typically sufficient to represent most continuous functions efficiently \citep{huang1997general,huang2003learning}, and a maximum of $2n+1$ nodes can capture complex patterns \citep{hecht1987kolmogorov}. In the current study, various experimental settings were explored, and hyper-parameters that yielded the best results were selected. {Experiments demonstrated that enhancing discriminator complexity (increased width or depth) generally improved feature learning, subsequently benefiting generator performance. However, an excessively complex discriminator negatively impacted generator learning, highlighting the necessity of careful hyperparameter tuning to maintain a proper balance.}

The learning rate in the AdamW gradient update significantly impacts the model's efficiency and effectiveness, as this hyperparameter affects the speed at which the model converges to an optimal solution. A learning rate that is too small may result in exceedingly slow learning or convergence to an undesired local minimum, while a learning rate that is too large can cause oscillations between different local minima. To address these challenges, cyclical learning rate (CLR) schedules \citep{smith2017cyclical} have been developed to dynamically adjust the learning rate throughout the training process. Unlike traditional fixed or monotonically decreasing learning rates, CLR oscillates the learning rate within predefined bounds, enabling the model to escape local minima and explore the optima more effectively. To ensure that the discriminator can learn more rapidly and provide more informative gradients to the generator, the discriminator can be assigned a higher learning rate compared to the generator \citep{heusel2017gans}. The base values for the CLR were selected from the range $\{1,2,3,4,5,6,7,8,9,10\} \times 10^{-3}$ for the discriminator and $\{1,2,3,4,5,6,7,8,9,10\} \times 10^{-4}$ for the encoder, and the upper bound was set to five times the base value. Experiments were conducted to identify the optimal base values for different tasks.

To ensure the convergence of the algorithm, the IW-ELBO value for each mini-batch is recorded. Every 100 fitting iterations, the average IW-ELBO across these 100 mini-batches is calculated. If this average does not improve after 500 such calculations, the fitting process is terminated, and convergence is considered achieved.

In exploratory studies, although different optimization runs may produce varied estimates, some results can be shown to be equivalent after suitable transformation. The equivalence is checked by comparing the loading matrices from each run, following the process suggested by \citet{urban2021deep}. First, the factors are rotated using the Geomin oblique rotation method \citep{yates1987multivariate}. Next, any factor whose loadings sum to a negative value is reversed \citep{asparouhov2009exploratory} so that all factors point in the same direction. Then, one matrix is chosen as a reference while the columns in the other matrices are rearranged to reduce the mean-squared error between matching entries. After aligning the matrices, Tucker’s congruence coefficient is calculated \citep{lorenzo2006tucker}; if this value is above 0.98, the solutions are considered equivalent \citep{maccallum1999sample}. The same process of inversion and rearrangement is applied to factor correlation matrices, where both rows and columns are reordered for a direct comparison.

\section{Experiment Results of the Comparison between AVB and VAE}
This section describes the results of the empirical study and simulation studies comparing the IWAVB and the IWAE by \citet{urban2021deep}. IWAE was shown to be more efficient \citep{urban2021deep} than the popular MH-RM algorithm \citep{cai2010high}, while IWAVB achieved similar finite sample performance as IWAE in normal latent variables cases but outperformed it when handling more complex latent variables distribution. {The algorithms were implemented in PyTorch~2.6.0.} All experiments were conducted on a Kaggle computational environment equipped with an Intel(R) Xeon(R) CPU @ 2.00 GHz (4 vCPUs) and two Tesla T4 GPUs (each with 16 GB of memory).

\subsection{Simulation Study: Latent Variables Following Normal Distribution }
\subsubsection{Comparison with IWAE: Importance-Weighting Technique}
This simulation study compared the IWAVB and IWAE methods in terms of parameter recovery and computational efficiency. Because both methods employ importance weighting, comparisons were conducted using various numbers of IW samples. The evaluation was based on a confirmatory analysis of a graded response model with $P=5$ latent factors, $N=500$ respondents, $M=50$ items, and $C_j=5$ response categories for $j=1,\dots,M=50$. Each factor was assumed to load on ten items, and each item corresponded to only one factor. Additionally, all loading values were constrained to be positive. The true correlation matrix of the factors was sampled from the Lewandowski-Kurowicka-Joe (LKJ) distribution \citep{lewandowski2009generating}, a commonly used prior for positive definite symmetric matrices with unit diagonals in Bayesian hierarchical modeling.

The true latent variables were generated as 500 samples from a multivariate normal distribution with a zero mean and the true correlation matrix. For the item parameters, each item's true loading value was sampled from a log-normal distribution, obtained by exponentiating samples from $\mathcal{N}(0, 0.5)$. The true intercept parameters were generated from another multivariate normal distribution with a zero mean and a covariance matrix drawn from the LKJ distribution, then sorted into strictly increasing order. Using the conditional probabilities defined by \textbf{Equation (\ref{eq:grm_prob})}, a categorical distribution was employed to sample 100 different response matrices. Both the IWAVB and IWAE methods were fitted to these simulated response datasets and compared in terms of estimation accuracy, computation time, and approximate log-likelihood. The estimation accuracy of the true parameters, $\xi$, was assessed using the mean-square error (MSE):
\begin{equation}\label{eq:mse}
    \text{MSE}(\hat{\xi},\xi) = \frac{1}{100}\sum\limits_{a=1}^{100}(\hat{\xi}^{(a)}-\xi)^2,
\end{equation}
where $\hat{\xi}^{(a)}$ represents the estimate from replication $a$. Bias is also computed by removing the square for each summand in the \textbf{Equation (\ref{eq:mse})}.

After some experimentation for hyper-parameter tunning, the neural network configuration for IWAVB method was decided as one hidden layer with 128 neurons for the encoder and two hidden layers with 256 and 128 neurons for the discriminator. Meanwhile, the base learning rates for the IWAVB method were set to $\eta_1=0.01$ for the discriminator network and $\eta_2=0.001$ for the encoder network. The inference model in the IWAE method had one hidden layer with 100 neurons and the base learning rate was $\eta=0.001$.

\begin{figure}[!htb]
  \centering
  \includegraphics[width=\textwidth]{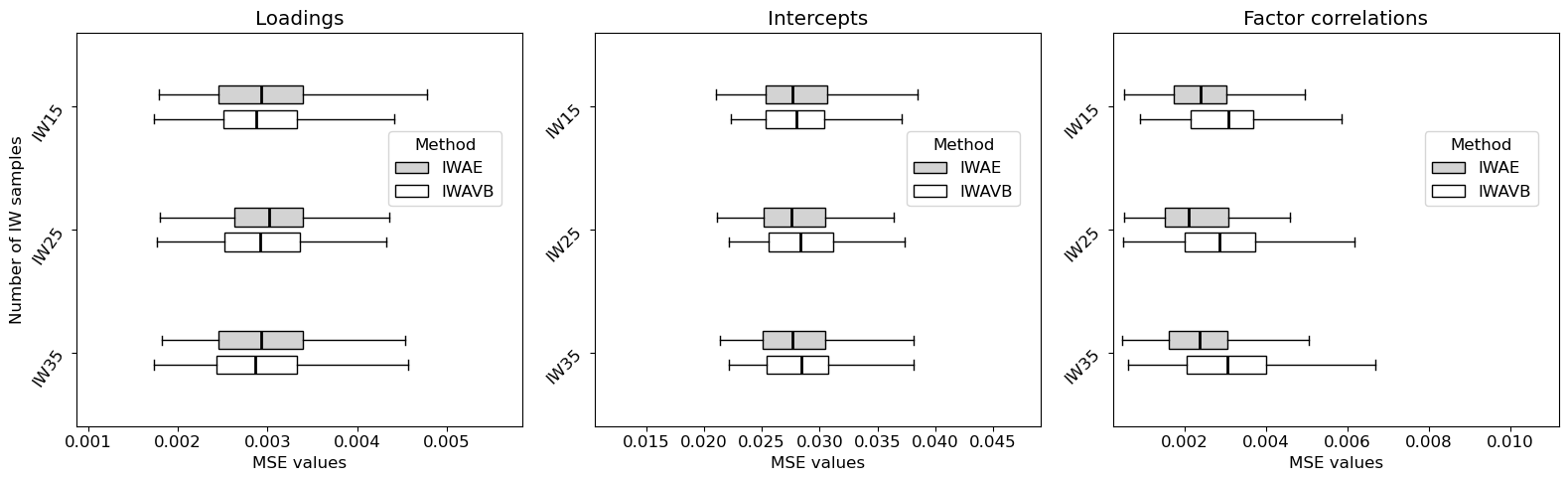}
  \includegraphics[width=\textwidth]{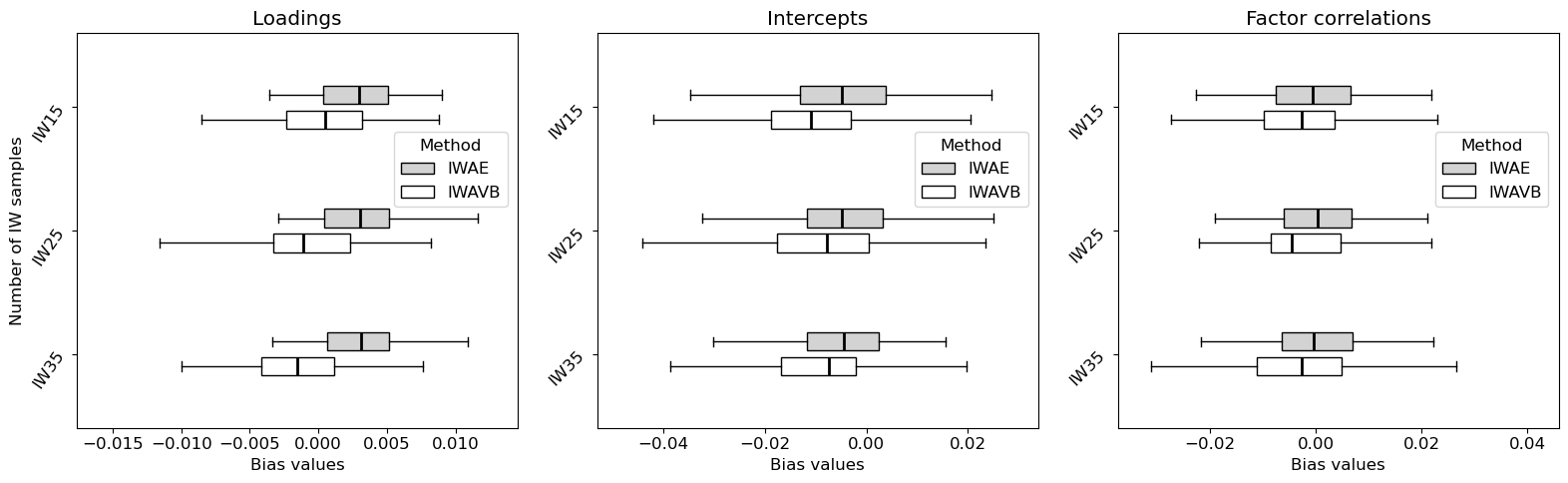}
  
  \caption{Parameter MSE and bias comparison for IWAVB and IWAE methods based on 100 replications of simulation with 5-dimensional latent variables following normal distribution.}
  \label{fig:mse_1}
\end{figure}

\textbf{Figure \ref{fig:mse_1}} shows box plots of the MSEs and bias for item parameters (loadings and intercepts), and factor correlations for the IWAVB and IWAE methods with 15, 25, or 35 importance-weighted (IW) samples. Both methods achieved low MSEs for estimating item loadings and factor correlations. While the MSE for item intercepts was relatively {high}, it still averaged below 0.03, which is satisfactory. Although the IWAE method slightly outperformed IWAVB in estimating item intercepts and factor correlations, the differences were minimal. Notably, IWAVB achieved a smaller MSE for loadings. {Bias values for both methods were consistently low, with loading estimates showing the lowest bias among the three parameter types.} Both methods showed marginal improvements in estimation accuracy as the number of IW samples increased.

\begin{figure}[htb]
\centering
\includegraphics[width=0.7\textwidth]{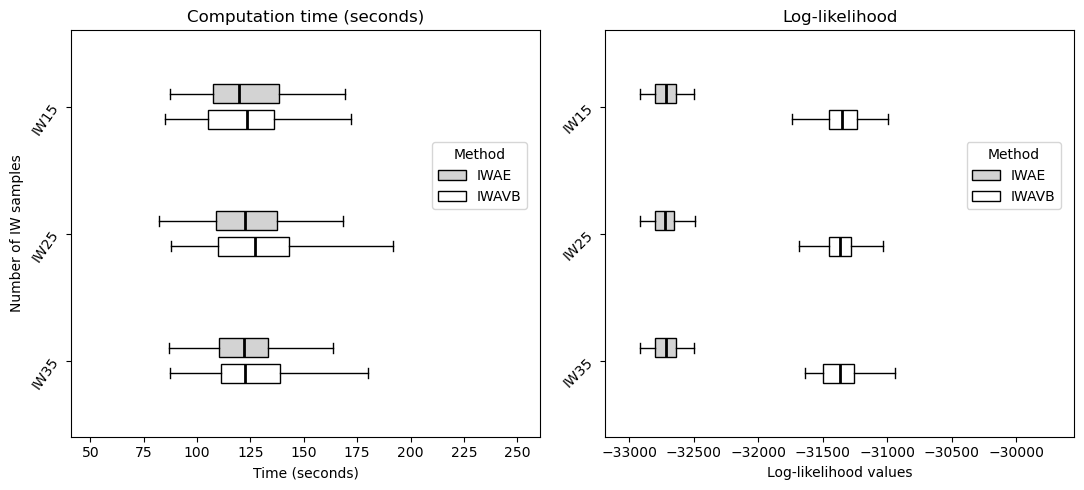}
\caption{Computation time and approximate log-likelihood comparison for IWAVB and IWAE methods across 100 replications of simulation with latent variables following normal distribution.}
\label{fig:time_ll_1}
\end{figure}

\textbf{Figure \ref{fig:time_ll_1}} presents box plots of computation time and approximate log-likelihood for the IWAVB and IWAE methods across 100 simulation replications. Both methods exhibited similar computation times, ranging from 110 to 140 seconds. Interestingly, IWAVB with 35 IW samples required less time than with 15 or 25 IW samples, suggesting that a larger number of IW samples might lead to better log-likelihood approximation and accelerate fitting. The approximate log-likelihood, computed on a 25\% holdout set randomly sampled from the response matrix (as in \textbf{Equation (\ref{eq:appro_ll})}), was significantly higher for the IWAVB method compared to the IWAE method.

\subsubsection{{Comparison with MH-RM: Higher Dimensionality and Larger Sample Size}}
To assess performance on more demanding problems, this study extended the comparison to include the established MH-RM algorithm. We used the MH-RM implementation from the \textit{mirt} (Version 1.44.0; \citealp{chalmers2012mirt}). Since its core functions are written in C++, it provides a relevant benchmark against the C++ backend of PyTorch. The comparison was based on two large-scale confirmatory analyses with the GRM, both using a sample size of $N=5000$ respondents: (1) a 7-dimensional model with $P=7$ latent factors and $M=70$ items; (2) a 10-dimensional model with $P=10$ latent factors and $M=100$ items. In both scenarios, each item had $C_j=5$ response categories. The factor structure remained simple, with each factor loading on ten unique items. The process for generating the true latent variables, item parameters, and the factor correlation matrix (from the LKJ distribution) was similar as the previous simulation.

Hyperparameter tuning indicated that the neural network configuration from the previous simulation remained effective for the 7-dimensional IWAVB model. For the 10-dimensional IWAVB model, one hidden layer with 170 neurons for the encoder and two hidden layers (400 and 200 neurons, respectively) for the discriminator provided optimal results. For IWAE, a single hidden layer with 100 neurons remained sufficient. Learning rates were unchanged from the prior simulations. The number of IW samples was set to 50 for IWAVB and IWAE methods to ensure stable performance.

\begin{figure}[!htb]
  \centering
  \includegraphics[width=\textwidth]{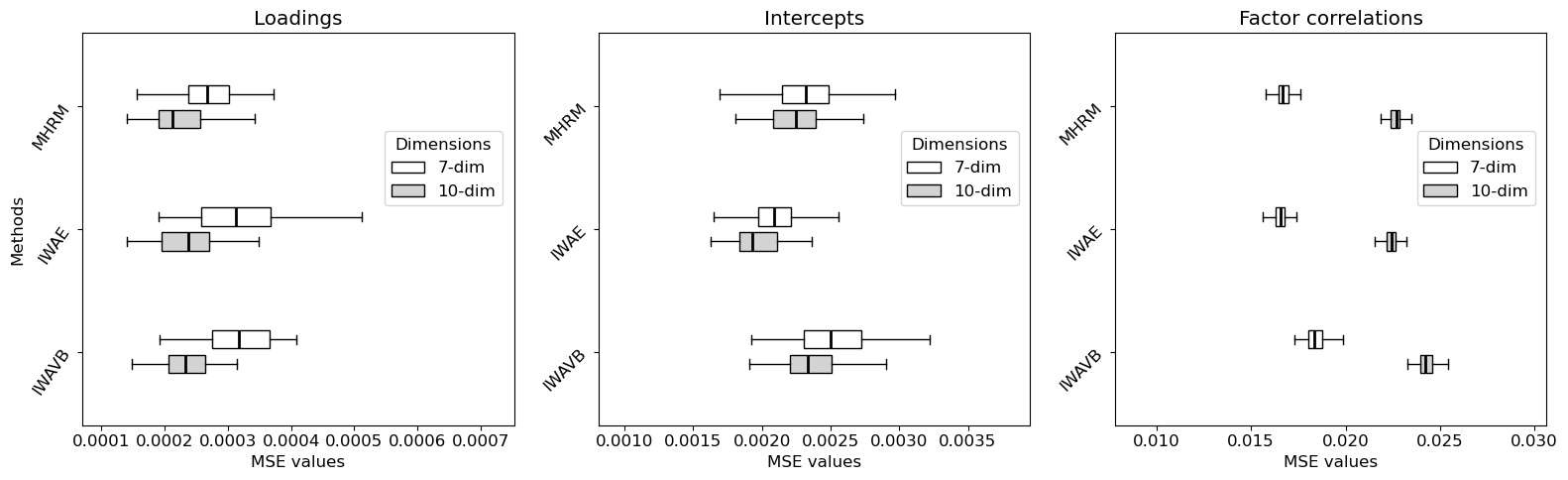}
  
  \caption{Parameter MSE comparison for IWAVB, IWAE and MHRM methods based on 100 replications of simulation with 7-dimensional and 10-dimensional latent variables following normal distribution.}
  \label{fig:mse_dim710}
\end{figure}

\textbf{Figure \ref{fig:mse_dim710}} presents box plots comparing MSEs for loadings, intercepts, and factor correlations across IWAVB, IWAE, and MH-RM. All three methods demonstrated low MSE values, with negligible differences between methods.

\textbf{Table \ref{tab:time_ll_dim710}} compares computation times among the three methods. Due to the lack of GPU support in \textit{mirt}, MH-RM required substantially more computation time, especially in higher-dimensional cases with larger sample sizes. IWAVB, while still slower than IWAE, significantly outperformed MH-RM in computational efficiency.
\begin{table}[H]
\centering
\caption{Computation time comparison for IWAE, IWAVB and MH-RM methods across 100 replications of simulation with 7-dimensional and 10-dimensional latent variables following normal distribution.}
\label{tab:time_ll_dim710}
\begin{tabular*}{\linewidth}{@{\extracolsep{\fill}} llll}
\toprule
 & \textbf{IWAE (7-dim)} & \textbf{IWAVB (7-dim)} & \textbf{MH-RM (7-dim)} \\
\midrule
\textbf{Mean computation time (seconds)} & 76.66 (SD=12.07) & 159.99 (SD=26.63) & 662.93 (SD=25.48) \\
\midrule
 & \textbf{IWAE (10-dim)} & \textbf{IWAVB (10-dim)} & \textbf{MH-RM (10-dim)} \\
\midrule
\textbf{Mean computation time (seconds)} & 80.64 (SD=14.00) & 179.03 (SD=29.85) & 1,666.54 (SD=43.71) \\
\bottomrule
\end{tabular*}

\vspace{0.5em}
{\footnotesize Note: SD = Standard Deviation.}
\end{table}

\subsection{Simulation Study: Latent Variables Following Multimodal Distribution}
\begin{figure}[htb]
\centering
\includegraphics[width=0.45\textwidth]{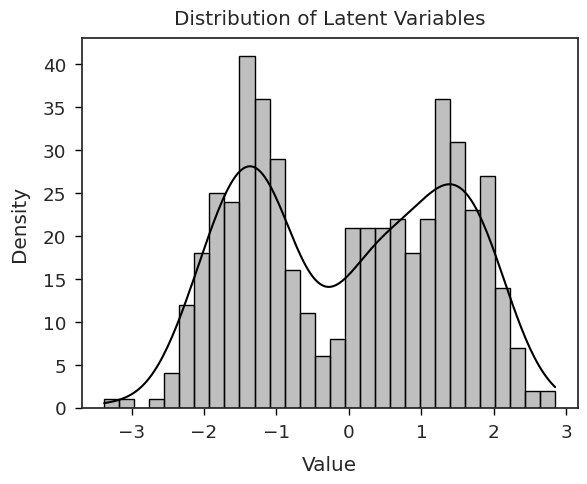}
\caption{Multimodal Distribution of Latent Variables.}
\label{fig:dist_mgrm}
\end{figure}

This simulation study aimed to evaluate a more general case where the latent variables follow a multimodal distribution. A graded response model was configured with $P=1$ latent factor, $N=500$ respondents, $M=20$ items, and $C_j=3$ response categories for $j=1,\dots,M=20$. The true latent variables were sampled from a mixture of three Gaussian distributions: $\mathcal{N}(-1.5, 0.5)$, $\mathcal{N}(0, 0.5)$, and $\mathcal{N}(1.5, 0.5)$, with respective weights of $0.4$, $0.2$, and $0.4$, as shown in \textbf{Figure \ref{fig:dist_mgrm}}. This resulted in a distribution with multiple modes, in contrast to the unimodal structure of a normal distribution. For the item parameters, 20 true loading values were generated from a log-normal distribution with a mean of zero and variance of $0.5$. True intercepts were sampled using a similar procedure as described in the previous simulation study. Additionally, a three-dimensional case was examined by configuring a GRM with $P=3$ latent factor, $N=5000$ respondents, $M=45$ items, and $C_j=3$ per item. Each item also corresponded to only one factor. In this instance, the true latent variables were generated from a mixture of three trivariate Gaussian distributions with mean vectors taking on the values $-1.5$, $0$, and $1.5$, and with a fixed correlation matrix sampled from the LKJ distribution. The true loading and intercept values were generated using analogous procedures as described previously.

\begin{figure}[!htb]
    \centering
    \includegraphics[width=\textwidth]{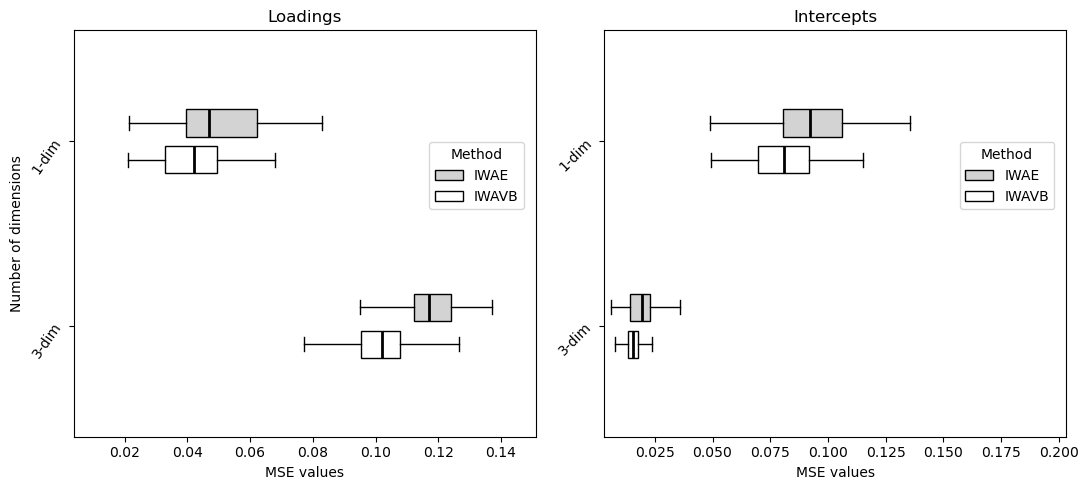}
    \caption{Parameter MSE comparison for IWAVB and IWAE methods based on 100 replications of simulation with latent variables following multimodal distribution.}
    \label{fig:mse_mgrm}
\end{figure}

We used the following base learning rates for the discriminator and encoder in the IWAVB method: $(\eta_1=0.003,\eta_2=0.0002)$ for $P=1$ and $(\eta_1=0.002,\eta_2=0.0005)$ for $P=3$, while for the inference model (one hidden layer with 100 neurons) in the IWAE method, both cases utilized $\eta=0.0005$. The configuration of neural networks was same as the first study. The number of IW samples was set to $R=30$ for both methods. Using these configurations, \textbf{Figure \ref{fig:mse_mgrm}} presents box plots comparing the MSEs for item loadings and intercepts estimated by the IWAVB and IWAE methods. The IWAVB method consistently achieved lower MSEs across both parameter estimates compared to IWAE.

\textbf{Table \ref{tab:time_ll_mgrm}} illustrates the computation time and approximate log-likelihood comparison for two methods in this simulation study. Although the IWAVB method required more computational time, it consistently achieved a higher approximate log-likelihood compared to the IWAE method.
\begin{table}[H]
\centering
\caption{Computation time and approximate log-likelihood comparison for IWAE and IWAVB methods across 100 replications of simulation with latent variables following multimodal distribution.}
\label{tab:time_ll_mgrm}
\begin{tabular*}{\linewidth}{@{\extracolsep{\fill}} lll}
\toprule
 & \textbf{IWAE (1-dim)} & \textbf{IWAVB (1-dim)} \\
\midrule
\textbf{Mean computation time (seconds)} & 162.01 (SD=25.19) & 181.03 (SD=22.09) \\
\textbf{Mean log-likelihood} & -5,233.35 (SD=80.38) & -5,129.92 (SD=102.19) \\
\midrule
 & \textbf{IWAE (3-dim)} & \textbf{IWAVB (3-dim)} \\
\midrule
\textbf{Mean computation time (seconds)} & 217.42 (SD=42.16) & 255.93 (SD=45.99) \\
\textbf{Mean log-likelihood} & -125,398.75 (SD=325.64) & -111,723.19 (SD=537.28) \\
\bottomrule
\end{tabular*}

\vspace{0.5em}
{\footnotesize Note: SD = Standard Deviation.}
\end{table}

\subsection{Empirical Study: Big-Five Personality Questionnaire}
In this empirical study, the performance of IWAVB method was evaluated in comparison to IWAE approach on a large-scale Big Five personality dataset. This dataset, included in \citet{urban2021deep}'s study, consists of responses to 
 \citet{goldberg1992development}'s 50-item Big-Five Factor Marker (FFM) questionnaire from the International Personality Item Pool (IPIP; \citealp{goldberg2006international}). The dataset was designed to measure five key dimensions of personality: 1. Extroversion; 2. Emotional stability; 3. Agreeableness; 4. Conscientiousness; 5. Openness. Each factor was assessed using ten items rated on a five-point scale, ranging from ``Disagree'' to ``Agree.'' The dataset is notable for its large scale, with over one million individual item responses, which makes it a challenging yet valuable benchmark for testing the efficiency and accuracy of high-dimensional exploratory factor analysis methods. After data pre-processing, the final sample size was $N=515,707$ responses.

To approximate the true log-likelihood for performance evaluation, \citet{urban2021deep} randomly sampled a subset of 2.5\% of respondents and created a holdout set, denoted as $\Omega$, using their responses. The model was trained on the remaining item responses, excluding the holdout set, and the fitted parameters were represented as $\hat{\bm{\theta}}$ and $\hat{\bm{\phi}}$. According to \textbf{Equation (\ref{eq:iw-elbo})}, the approximate log-likelihood for the holdout set with $R=5000$ IW samples could be calculated as:
\begin{equation}\label{eq:appro_ll}
    \sum\limits_{i\in\Omega}\left[\log\frac{1}{5000}\sum\limits_{r=1}^{5000}\frac{p_{\hat{\bm{\theta}}}(\bm{z}_{i,r},\bm{x}_i)}{q_{\hat{\bm{\phi}}}(\bm{z}_{i,r}\mid\bm{x}_i)}\right].
\end{equation}

\begin{figure}[htb]
\centering
\includegraphics[width=0.7\textwidth]{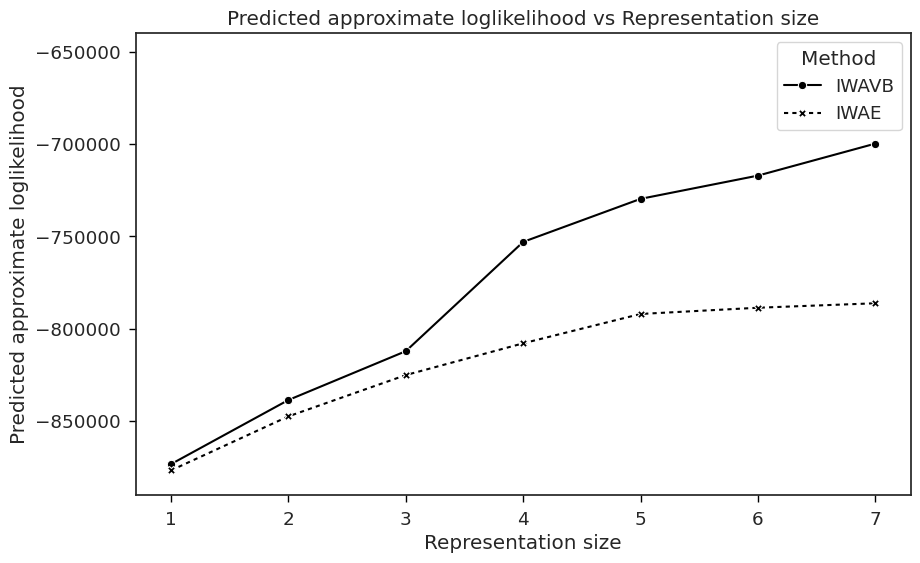}
\caption{Scree plot of predicted approximate log-likelihood with different number of latent factors.}
\label{fig:ll_vs_repr}
\end{figure}

\textbf{Figure \ref{fig:ll_vs_repr}} presents a scree plot of the predicted approximate log-likelihood for varying numbers of latent factors $P \in \{1, 2, 3, 4, 5, 6, 7\}$. The plot reveals an ``elbow'' at $P=5$, indicating that five latent factors were sufficient to capture the majority of the correlation among item responses. With several preliminary experiments to reveal the appropriate network sizes for the IWAVB method, we configured the encoder network with one hidden layer containing 256 neurons and the discriminator network with two hidden layers consisting of 512 and 256 neurons, respectively. To achieve sufficiently accurate $\bm{\theta}$ estimates, the size of IW samples was set to $R=25$ for the IWAVB and IWAE methods. The base learning rates for the IWAVB method were set to $\eta_1=0.01$ for the discriminator and $\eta_2=0.001$ for the encoder, while the base learning rate for the inference model (a single hidden layer with 130 neurons) in the IWAE method was set to $\eta=0.001$. The mini-batch size is set to $256$. Results demonstrated that the IWAVB method consistently achieved higher approximate log-likelihoods compared to the IWAE method. Furthermore, the improvement in log-likelihood was more significant for the IWAVB method as the number of latent factors increased.

For the latent factor model with $P=5$ factors, we compared the performance of two methods. To assess the replicability of the estimation across different random seeds, the fitting process was repeated 100 times on the complete data set. {Among these repetitions, the model with the highest approximate log-likelihood was chosen as the \textit{reference model}.}

\begin{figure}[htb]
    \centering
    \subfigure[IWAE]{
        \includegraphics[width=0.48\textwidth]{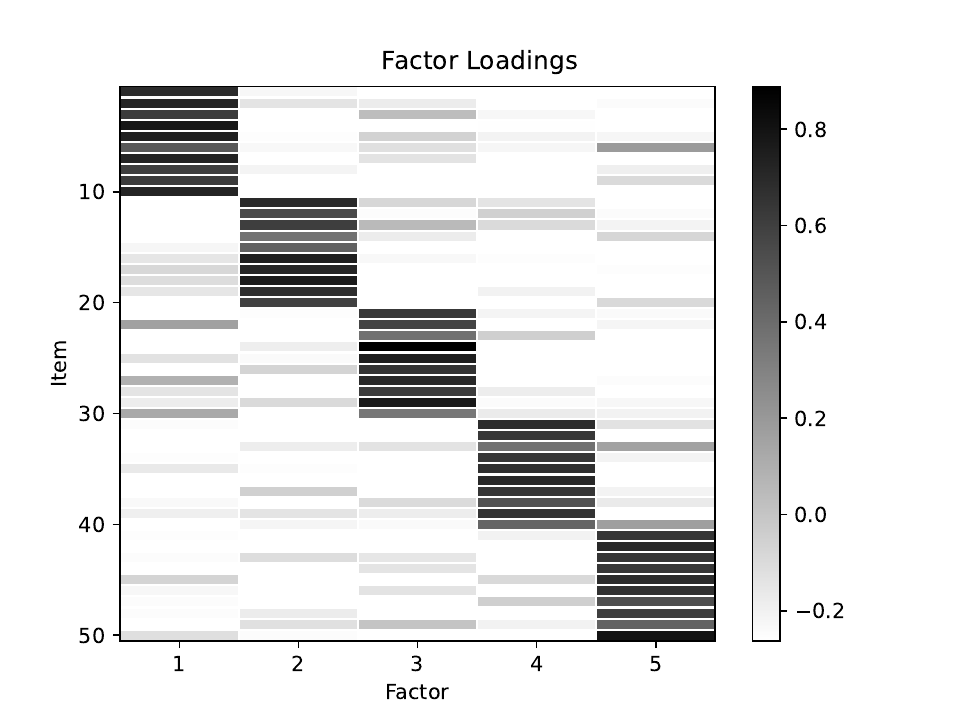}
        \label{fig:load_vae}}
    \subfigure[IWAVB]{
        \includegraphics[width=0.48\textwidth]{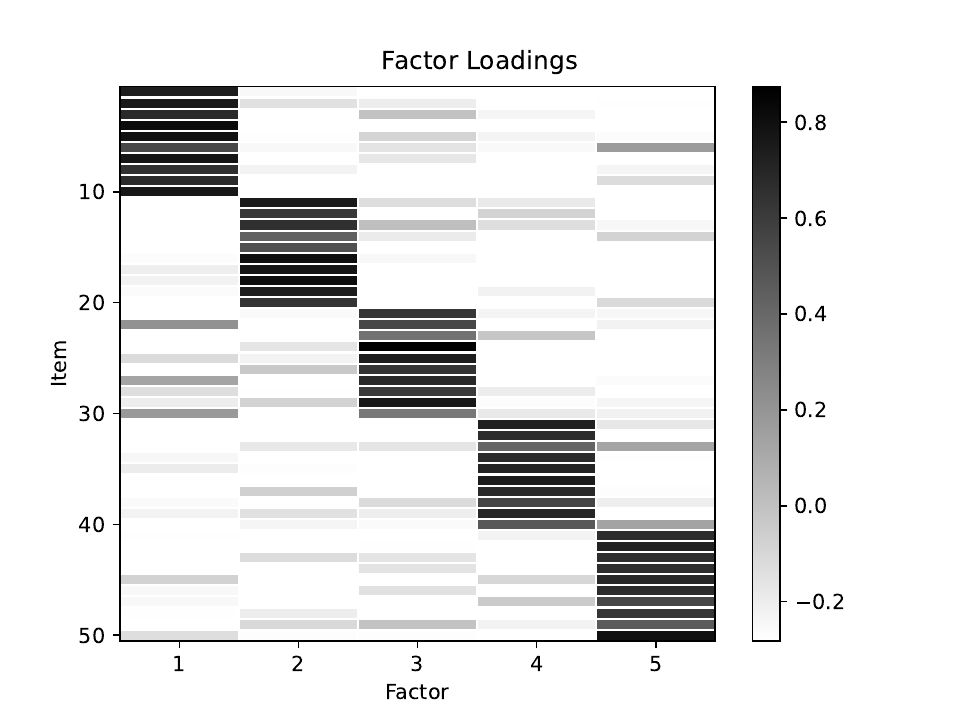}
        \label{fig:load_avb}}

    \caption{Heat map of factor loadings estimated by IWAE and IWAVB for IPIP-FFM items.}
    \label{fig:load_avb_vae}
\end{figure}

To enhance the interpretability of the loading matrix, we applied the Geomin oblique rotation method \citep{yates1987multivariate} to the loading matrices produced by both algorithms. This method allows factors to correlate, which typically captures the underlying structure more faithfully when the factors are not entirely independent. As shown in \textbf{Figure \ref{fig:load_avb_vae}}, the heat maps of the Geomin-rotated loadings from the reference models of both methods reveal a highly similar and theoretically expected five-factor configuration.

{Let $\xi_{ref}$ denote one of the parameter estimates obtained from the reference model}, and the approximate root-mean-square error (RMSE) across repetitions is defined as follows:
\begin{equation}\label{eq:rmse}
    \text{RMSE}(\hat{\xi},\xi_{ref}) = \sqrt{\frac{1}{99}\sum\limits_{a=1,\,\hat{\xi}^{(a)}\neq\xi_{ref}}^{100}(\hat{\xi}^{(a)}-\xi_{ref})^2},
\end{equation}
where $\hat{\xi}^{(a)}$ represents the estimate from replication $a$. \textbf{Table \ref{tab:emp_results}} presents the RMSE for the estimates of the loadings, intercept, and covariance matrix, calculated relative to the reference models. The results indicate that parameter estimation was quite stable for both methods. Moreover, the table also reports the mean computation time, showing that the IWAVB method requires more time. This increased computation time is attributed to the inference model in IWAVB having a larger number of neurons and an additional discrimination network, which, although leading to a more general model, results in a more complex and computationally demanding structure.
\begin{table}[H]
\centering
\caption{Performance metrics for IPIP-FFM data set using IWAVB and IWAE methods.}
\label{tab:emp_results}
\begin{tabular*}{\linewidth}{@{\extracolsep{\fill}} lll}
\toprule
 & \textbf{Mean loadings RMSE} & \textbf{Mean intercepts RMSE} \\
\midrule
\textbf{IWAVB} & 0.017 (SD = 0.016) & 0.092 (SD = 0.057) \\
\textbf{IWAE} & 0.019 (SD = 0.012) & 0.052 (SD = 0.045) \\
\midrule
 & \textbf{Mean covariance matrix RMSE} & \textbf{Mean computation time (seconds)} \\
\midrule
\textbf{IWAVB} & 0.085 (SD = 0.010) & 243.89 (SD = 33.67) \\
\textbf{IWAE} & 0.078 (SD = 0.026) & 85.96 (SD = 16.75) \\
\bottomrule
\end{tabular*}

\vspace{0.5em}
{\footnotesize Note: SD = Standard Deviation.}
\end{table}

\section{Discussion}
This study introduces and applies an integration of VAEs and GANs to IFA, two prominent paradigms in representation learning, in the context of psychometrics. The introduced method, Adversarial Variational Bayes, is further refined by incorporating the importance-weighted technique, resulting in an innovative variational approach for IFA parameter estimation. IWAVB improves the Variational Autoencoder algorithm by removing the normal distribution constraint on the inference model. Instead, a discriminator network is trained to achieve a more expressive and flexible inference model. 

We conducted three comparative studies between IWAE and IWAVB to evaluate their performance. In the first simulation study using a normal distribution for latent variables, IWAVB achieved similarly low mean squared error and higher log-likelihood compared to IWAE, while demonstrating its higher efficiency in standard conditions than established MH-RM algorithm. Furthermore, in a more general and realistic scenario where the latent variables follow a multimodal distribution, IWAVB outperformed IWAE in the estimation of all parameters and log-likelihood, highlighting its robustness and superior performance in complex and real-world settings. The empirical study, based on the large-scale Big-Five personality factors dataset, yielded parameter estimate results comparable to the IWAE method while achieving higher model likelihood, indicating a better fit to the data. 

A critical reason behind IWAVB's improved performance in the multimodal setting is its flexibility in modeling various latent structures. Although IWAE employs importance weighting to refine the variational objective, it still relies on a more traditional VAE architecture that often assumes unimodal or near-unimodal latent distributions. By contrast, IWAVB employs adversarial training strategies while capitalizing on a flexible feedforward neural network to capture intricate probability densities. This adversarial component helps the inference mechanism explore multiple latent clusters more effectively.

Nonetheless, the proposed IWAVB method is not without limitations that need further improvement. {First, as with many GAN-based methods, training can be unstable. This instability arises from the adversarial process, in which the generator and discriminator compete rather than optimize a single objective. Consequently, issues such as mode collapse may occur, where an overly strong discriminator in early training restricts the generator's ability to explore the latent space \citep{mi2018probe}. Although adjusting hyperparameters or network architecture can mitigate these problems, our experiments show that IWAVB is highly sensitive to such changes, making the tuning process time-consuming.} Furthermore, exploring different neural network specification, such as altering the number or size of hidden layers, or integrating more advanced architectures such as attention, may prove beneficial for diverse tasks, yet there is no standardized IFA extension nor well-defined criterion to guide these extensions.

Despite these limitations, the concept of leveraging neural networks to represent complex probabilistic models remains promising in psychometrics and beyond. In particular, because the decoder currently relies on a simple sigmoid-activated linear function, more sophisticated neural architectures could improve parameter estimation and provide methodological foundations to extend IFA to accommodate multimodal data inputs, such as modeling image and texts \citep{cheng2019dirt}. Meanwhile, AVB provides a framework that bridges GANs with Marginal Maximum Likelihood, thereby enabling the adaptation of numerous GAN variants, such as Deep Convolutional GANs \citep{radford2015unsupervised}, Wasserstein GANs \citep{arjovsky2017wasserstein}, and CycleGANs \citep{zhu2017unpaired}, to potentially improve estimation in IFA and beyond.

Beyond the psychometrics literature, in computer science, IFA is increasingly applied to improve the interpretability of sophisticated neural networks. For example, \citet{cheng2019dirt} employed LSTM architectures to handle question text and multiple knowledge concepts for cognitive diagnosis, whereas \citet{yeung2019deep} merged IFA with a recurrent neural network (RNN) to develop a more transparent knowledge-tracing model. Although these studies emphasize predicting respondents’ future performance, rather than recovering parameters, they underscore the growing intersection of IFA with advanced neural methods and underscore the crucial potential that flexible inference methods, such as AVB, can play in real-world applications.

In summary, IWAVB offers a novel pathway to more flexible and powerful inference within Item Factor Analysis, outstripping existing methods when confronted with complex latent distributions. While challenges remain, its ability to capture richer latent structures suggests that adversarially assisted, importance-weighted methods will drive further theoretical and practical advancements in parameter estimation and model extensions for IFA. The enhanced inference model, powered by GANs, provides the methodological foundation for IFA to be further developed to potentially model diverse and complex data, including text, images, and videos.

\section*{Appendix}\label{sec:appendix}
Considering $P_{l}\times P_{l-1}$ matrix $\bm{W}_0^{(l)}=[w_{p_l,p_{l-1},0}]$ and $P_{l}\times 1$ vector $\bm{b}_0^{(l)}=(b_{p_l,0})$, Kaiming initialization is defined as follows:
\begin{equation}\label{eq:kaiming}
    w_{p_l,p_{l-1},0}, b_{p_l,0} \sim \mathcal{U}\left(-\sqrt{\frac{3}{P_{l-1}}},\sqrt{\frac{3}{P_{l-1}}}\right),
\end{equation}
where $\mathcal{U}(a,b)$ represents the uniform distribution with lower bound $a$ and upper bound $b$. 

For item $j=1,\dots,M$, the $P\times 1$ vector of loadings $\bm{\beta}_{j,0}=(\beta_{p,j,0})$ and the $(C_j-1)\times 1$ vector of intercepts $\bm{\alpha}_{j,0}=(\alpha_{k,j,0})$ are initialized by: for $p=1,\dots,P$ and $k=1,\dots,C_j-1$,
\begin{equation}\label{eq:xavier}
    \beta_{p,j,0},\alpha_{k,j,0} \sim \mathcal{U}(-\sqrt{\frac{2}{P+M}},\sqrt{\frac{2}{P+M}}).
\end{equation}

AdamW separates the weight decay term from the gradient updates, which enables better generalization in deep learning models because it explicitly penalizes large weights through a regularization term without conflating it with the optimization step. The parameter update rule in AdamW includes momentum and second-order moment estimation, and is expressed as follows:
\begin{align}
    &\text{Momentum update: }\bm{m}_t = \beta_1 \bm{m}_{t-1} + (1 - \beta_1) \nabla_{\bm{\omega}}\mathcal{L}(\bm{\omega}),\\
    &\text{Second moment update: }\bm{v}_t = \beta_2 \bm{v}_{t-1} + (1 - \beta_2) (\nabla_{\bm{\omega}}\mathcal{L}(\bm{\omega}))^2,\\
    &\text{Weight decay adjustment: }\bm{\omega}_{t+1} = \bm{\omega}_t - \eta \frac{\bm{m}_t}{\sqrt{\bm{v}_t} + \epsilon} - \eta \lambda \bm{\omega}_t,
\end{align}
where $\bm{m}_t$ and $\bm{v}_t$ are the first and second moment estimates, $\beta_1$ and $\beta_2$ are exponential decay rates for the first and second moment estimates, $\lambda$ is the weight decay coefficient, and $\epsilon$ is a small constant for numerical stability. The hyperparameters are set to the default values ($\beta_1=0.9, \beta_2=0.999, \lambda=0.01, \epsilon=1\times10^{-8}$) in this study, as they are practically recommended \citep{loshchilov2017decoupled,reddi2019convergence}. 

\bibliography{cite}
\end{document}